\documentclass{article}

    \PassOptionsToPackage{numbers, compress}{natbib}

 \usepackage[main, final]{neurips_2026}

\usepackage[utf8]{inputenc} 
\usepackage[T1]{fontenc}    
\usepackage{hyperref}       
\usepackage{url}            
\usepackage{booktabs}       
\usepackage{amsfonts}       
\usepackage{nicefrac}       
\usepackage{microtype}      
\usepackage{xcolor}         
\usepackage{graphicx} 
\usepackage{amsmath}
\usepackage{multirow}
\usepackage{subcaption}
\usepackage{pifont}
\usepackage{colortbl}
\usepackage{wrapfig}

\newcommand{\etc}{\textit{etc}}
\newcommand{\eg}{\textit{e.g.}}

\newcommand{\ie}{\emph{i.e.}}

\definecolor{tabfirst}{rgb}{1, 0.7, 0.7}
\definecolor{tabsecond}{rgb}{1, 0.85, 0.7}
\definecolor{tabthird}{rgb}{1, 1, 0.7}

\title{TurboVGGT: Fast Visual Geometry Reconstruction with Adaptive Alternating Attention}

\author{%
  David Huang$^{1,2,}$\thanks{David Huang contributed to this work during an internship at Huawei Canada.}~$^{,}$\thanks{Equal contribution.} \quad Guile Wu$^{1,}$\footnotemark[2] \quad Chengjie Huang$^1$ \quad Bingbing Liu$^3$ \quad Dongfeng Bai$^1$\\
  $^{1}$Huawei Noah's Ark Lab \quad $^{2}$University of Toronto \quad $^{3}$Foundation Model Department, Huawei\\
  \texttt{dawae.huang@mail.utoronto.ca, guile.wu@outlook.com, chengjie.huang@outlook.com,} \\
  \texttt{liu.bingbing@huawei.com, baidongfeng@huawei.com} \\
  \texttt{\url{https://turbovggt.github.io/}}\\
}

\begin{document}

\maketitle

\begin{figure}[h]
    \centering
    \includegraphics[width=0.99\textwidth]{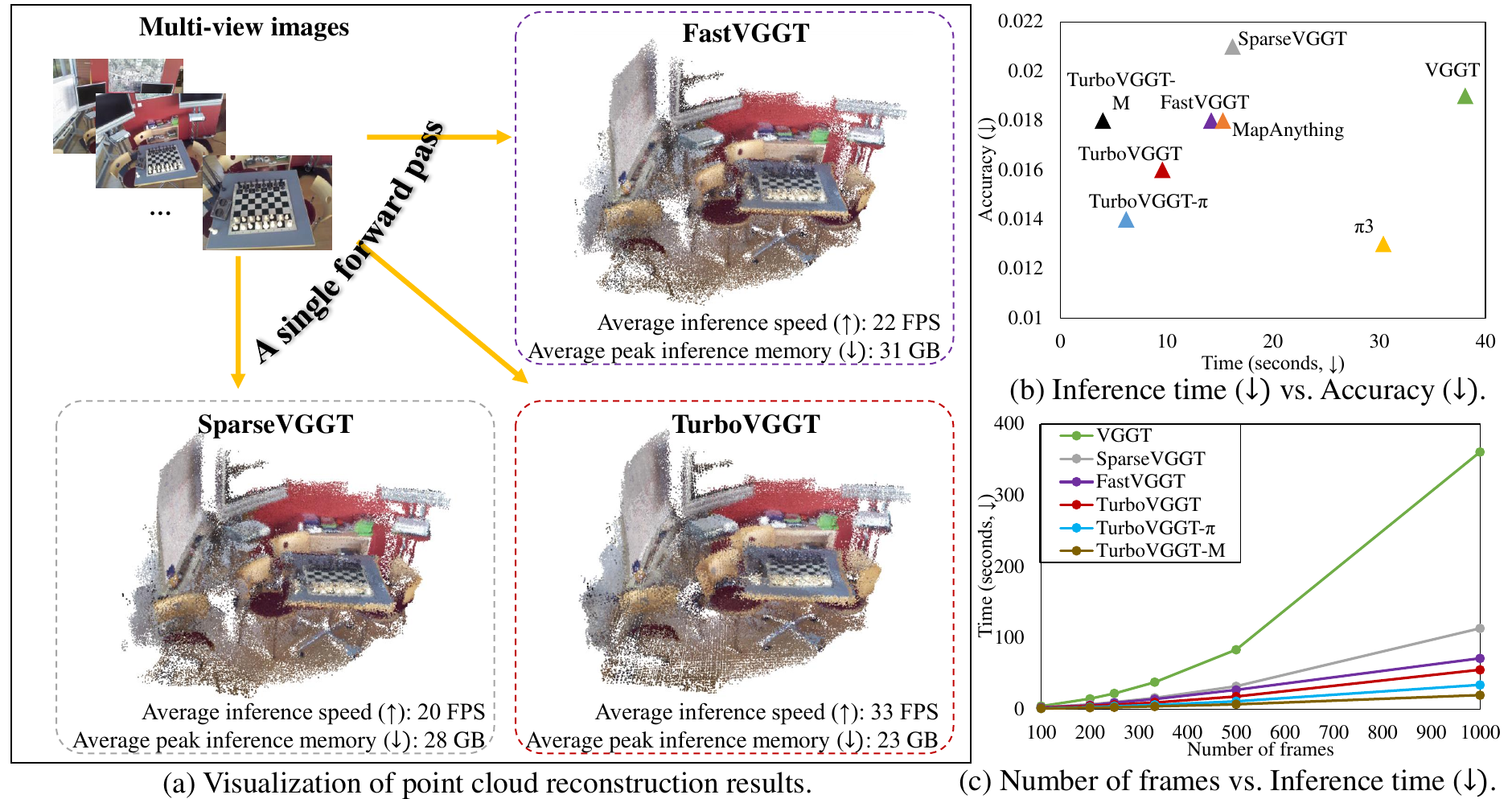}
    \caption{An illustration of our TurboVGGT for fast multi-view 3D reconstruction.
    (a) Visualization of point cloud reconstruction results.
    TurboVGGT significantly improves computational efficiency in terms of inference time and peak GPU memory usage.
    (b) Comparison of inference time vs. point cloud accuracy on 7-Scenes (stride 3).
    TurboVGGT achieves a better trade-off between accuracy and efficiency compared to state-of-the-art methods.
    (c) Comparison of number of frames vs. inference time on 7-Scenes.
    TurboVGGT scales more efficiently with the number of input frames.
    TurboVGGT/TurboVGGT-$\pi$/TurboVGGT-M achieve 7$\times$/11$\times$/18$\times$ speedup compared to VGGT for processing input sequences of 1000 frames.
    }
    \label{fig:illustration}
\end{figure}

\begin{abstract}
Recent feed-forward 3D reconstruction methods, such as visual geometry transformers, have substantially advanced the traditional per-scene optimization paradigm by enabling effective multi-view reconstruction in a single forward pass.
However, most existing methods struggle to achieve a balance between reconstruction quality and computational efficiency, which limits their scalability and efficiency.
Although some efficient visual geometry transformers have recently emerged, they typically use the same sparsity ratio across layers and frames and lack mechanisms to adaptively learn representative tokens to capture global relationships, leading to suboptimal performance.
In this work, we propose TurboVGGT, a novel approach that employs an efficient visual geometry transformer with adaptive alternating attention for fast multi-view 3D reconstruction.
Specifically, TurboVGGT employs an end-to-end trainable framework that leverages adaptive sparse global attention guided by adaptive sparsity selection to capture global relationships across frames and uses frame attention to aggregate local details within each frame.
In the adaptive sparse global attention, TurboVGGT adaptively learns representative tokens with varying sparsity levels for global geometry modeling, considering that token importance varies across frames, different attention layers operate on tokens at different levels of abstraction, and global dependencies rely on structurally informative regions.
Extensive experiments on multiple 3D reconstruction benchmarks demonstrate that TurboVGGT achieves fast multi-view reconstruction while maintaining competitive reconstruction quality compared with state-of-the-art methods.
\end{abstract}

\section{Introduction}

Multi-view 3D reconstruction is a long-standing problem in computer vision.
It aims to reconstruct the 3D geometry of a scene from a set of 2D images captured from different views.
It has been widely used in various applications, such as robotics, autonomous driving, and virtual reality.
Traditional methods, such as Structure-from-Motion (SfM)~\cite{schonberger2016structure} and Multi-View Stereo (MVS)~\cite{schonberger2016pixelwise}, typically employ a multi-stage paradigm to estimate camera poses and reconstruct 3D geometry.
This paradigm is effective in producing accurate reconstructions but suffers from high computational costs.

Recently, learning-based 3D reconstruction~\cite{wang2024dust3r,wang2025vggt,wang2025continuous,keetha2026mapanything} has emerged as a promising alternative to the traditional paradigm, shifting from per-scene iterative optimization to direct inference of 3D geometry from images using end-to-end neural networks.
In particular, the emergence of visual geometry transformer based feed-forward reconstruction methods, such as VGGT~\cite{wang2025vggt}, MapAnything~\cite{keetha2026mapanything}, and $\pi^3$~\cite{wang2025pi}, have revolutionized multi-view 3D reconstruction, enabling efficient and robust estimation of camera poses, depth maps, and point maps in a single forward pass.
Despite the significant progress, these methods still face challenges in achieving a balance between reconstruction quality and computational efficiency.
This limits their scalability and hinders their application in time-sensitive scenarios.
More recently, several studies~\cite{shen2026fastvggt,wang2025flashvggt,sun2025avggt,wang2025faster} have revealed that the high computational cost of visual geometry transformers primarily lies in the dense global token interactions within global attention layers, which involve quadratic complexity with respect to the number of input tokens.
To resolve this problem, existing methods typically use token merging, sparse attention or token subsampling strategies to reduce the number of tokens involved in the global attention.
Although these methods can accelerate visual geometry transformers for 3D reconstruction, they still have some limitations.
First, they typically employ the same sparsity ratio across layers and frames.
This may be suboptimal because token importance can vary across frames and attention layers operate on tokens at different levels of abstraction.
Second, they lack mechanisms to adaptively learn representative tokens that can effectively capture global relationships.
Since global dependencies often rely on structurally informative regions, learning representative tokens can facilitate global geometry modeling and reduce redundant computations.

To address these limitations, in this work, we propose a novel approach named TurboVGGT, which improves the efficiency of visual geometry transformers for fast multi-view 3D reconstruction.
The key idea of TurboVGGT is to construct a visual geometry transformer with adaptive alternating attention blocks, which leverage adaptive sparse global attention guided by adaptive sparsity selection to capture global relationships and use frame attention to aggregate local details within each frame.
Specifically, considering that token importance can vary across frames and attention layers operate on tokens at different levels of abstraction, TurboVGGT employs a gating network to adaptively assign different sparsity ratios for different frames across global attention layers.
Then, in each branch corresponding to a specific sparsity level, TurboVGGT employs adaptive sparse global attention to learn compressed representative tokens for global geometry modeling.
This enables TurboVGGT to effectively learn representative tokens to capture global relationships while reducing redundant computations.
In addition, following existing works~\cite{wang2025vggt}, TurboVGGT also incorporates frame attention to aggregate local details within each frame.
We conduct extensive experiments on multiple 3D reconstruction benchmarks to evaluate the proposed TurboVGGT.
Experimental results demonstrate the superiority of TurboVGGT over state-of-the-art methods in achieving fast multi-view 3D reconstruction while maintaining high reconstruction quality (as illustrated in Figure~\ref{fig:illustration}).

In summary, our \textbf{contributions} are:
\begin{itemize}
    \item We present a novel visual geometry transformer with adaptive alternating attention blocks for fast multi-view 3D reconstruction.
    \item We propose an adaptive sparsity selection mechanism for visual geometry transformers, which adaptively selects different sparsity ratios for different frames across different layers.
    \item We propose adaptive sparse global attention for visual geometry transformers, which learns representative tokens to model global geometry relationships.
    \item We conduct extensive experiments on multiple 3D reconstruction benchmarks and demonstrate the superiority of the proposed approach over state-of-the-art methods.
\end{itemize}

\section{Related Work}

\subsection{Multi-View 3D Reconstruction}
Reconstructing 3D scenes from multi-view images is a fundamental problem in computer vision.
The traditional paradigm consists of multiple stages, such as feature extraction and matching, camera pose estimation, triangulation, and bundle adjustment.
Typically, Structure-from-Motion~\cite{schonberger2016structure,pan2024global} is employed to estimate camera parameters and reconstruct sparse point clouds from multi-view images, and Multi-View Stereo~\cite{schonberger2016pixelwise,furukawa2015multi} is used to reconstruct dense geometry of 3D scenes.
Although these methods can achieve high-quality reconstruction, they rely on per-scene iterative optimization, which is computationally expensive and time-consuming.
Recently, learning-based methods have been proposed to directly predict 3D geometry from multi-view images in end-to-end neural networks~\cite{wang2024dust3r,wang2025vggt,leroy2024grounding,keetha2026mapanything,yang2025fast3r,wang2025pi}.
DUSt3R~\cite{wang2024dust3r} is a pioneering work that employs transformer networks to directly predict dense 3D geometry from image pairs, but it requires costly global alignment when extending to large scenes due to the reliance on pairwise view processing.
Subsequent works, such as MASt3R~\cite{leroy2024grounding}, CUT3R~\cite{wang2025continuous}, Fast3R~\cite{yang2025fast3r}, DDUSt3R~\cite{han2025enhancing}, and VGGT~\cite{wang2025vggt}, have presented various strategies to resolve the limitations of DUSt3R for different problem settings, such as large scenes, long sequences, and dynamic scenes.
Our work belongs to the learning-based paradigm, and we focus on improving the efficiency of visual geometry transformers for fast multi-view 3D reconstruction, which is a critical problem but has not been well explored in existing works.

\subsection{Visual Geometry Transformers}
Visual geometry transformers~\cite{wang2025vggt,keetha2026mapanything,wang2025pi,zhuo2026streaming} have recently emerged as powerful models for feed-forward multi-view 3D reconstruction.
They can directly predict camera intrinsics, camera extrinsics, depth maps, point maps, and other geometric properties from multi-view images in a single forward pass.
VGGT~\cite{wang2025vggt} is a pioneering work that constructs a visual geometry transformer and optimizes it with multi-task learning and large-scale datasets.
$\pi^3$~\cite{wang2025pi} is a subsequent work that introduces a fully permutation-equivariant architecture to eliminate the reliance on a fixed reference view in VGGT.
MapAnything~\cite{keetha2026mapanything} proposes a unified feed-forward transformer that supports various problem settings with different input and output configurations.
Despite their impressive performance, they still face challenges in computational efficiency.
This limits their scalability and efficiency in time-sensitive application scenarios.
Although there have been some recent efforts~\cite{shen2026fastvggt,wang2025flashvggt,sun2025avggt,wang2025faster} to accelerate visual geometry transformers, they still have some limitations, such as relying on a fixed sparsity ratio for all layers and frames and lacking mechanisms to adaptively learn representative tokens for global geometry modeling.
To address these limitations, our work proposes a novel visual geometry transformer with adaptive alternating attention blocks, which can adaptively select different sparsity ratios for different frames and learn representative tokens to model global geometry while reducing redundant computations, thus improving the efficiency of feed-forward 3D reconstruction.

\subsection{Efficient Vision Transformers}
Vision transformers~\cite{dosovitskiy2021an} have achieved incredible success in computer vision.
To make vision transformers more efficient, various strategies have been used, such as FlashAttention~\cite{dao2022flashattention,dao2023flashattention}, sparse attention~\cite{wei2023sparsifiner,zhang2025spargeattention}, token merging~\cite{bolya2023token,zeng2022not}, and low-rank approximation~\cite{jaegle2021perceiver,han2024agent}.
Despite their efficiency, most efficient vision transformers are designed for general 2D vision tasks.
In the context of multi-view 3D reconstruction, designing an efficient visual transformer remains an open question, as it requires modeling complex global geometric relationships across views and local geometric details within views.
With the success of visual geometry transformers for multi-view 3D reconstruction, there have been some recent efforts employing token merging~\cite{shen2026fastvggt}, block-sparse attention~\cite{wang2025faster}, subsampling global attention~\cite{sun2025avggt}, and compressed descriptors~\cite{wang2025flashvggt} to accelerate visual geometry transformers.
Our work belongs to this line of research, but unlike existing works, we propose to accelerate visual geometry transformers with adaptive alternating attention blocks.
In addition, DiffRate~\cite{chen2023diffrate} learns differentiable layer-wise compression rates for conventional 2D vision transformers, while GitMerge3D~\cite{tran2026many} performs graph-based token merging for transformers operating on explicit 3D point clouds.
In contrast, TurboVGGT targets multi-view image-based geometry transformers and adaptively allocates sparsity across frames and layers while learning representative tokens for cross-view geometry aggregation.

\section{Methodology}

\begin{figure}[t]
    \centering
    \includegraphics[width=0.99\textwidth]{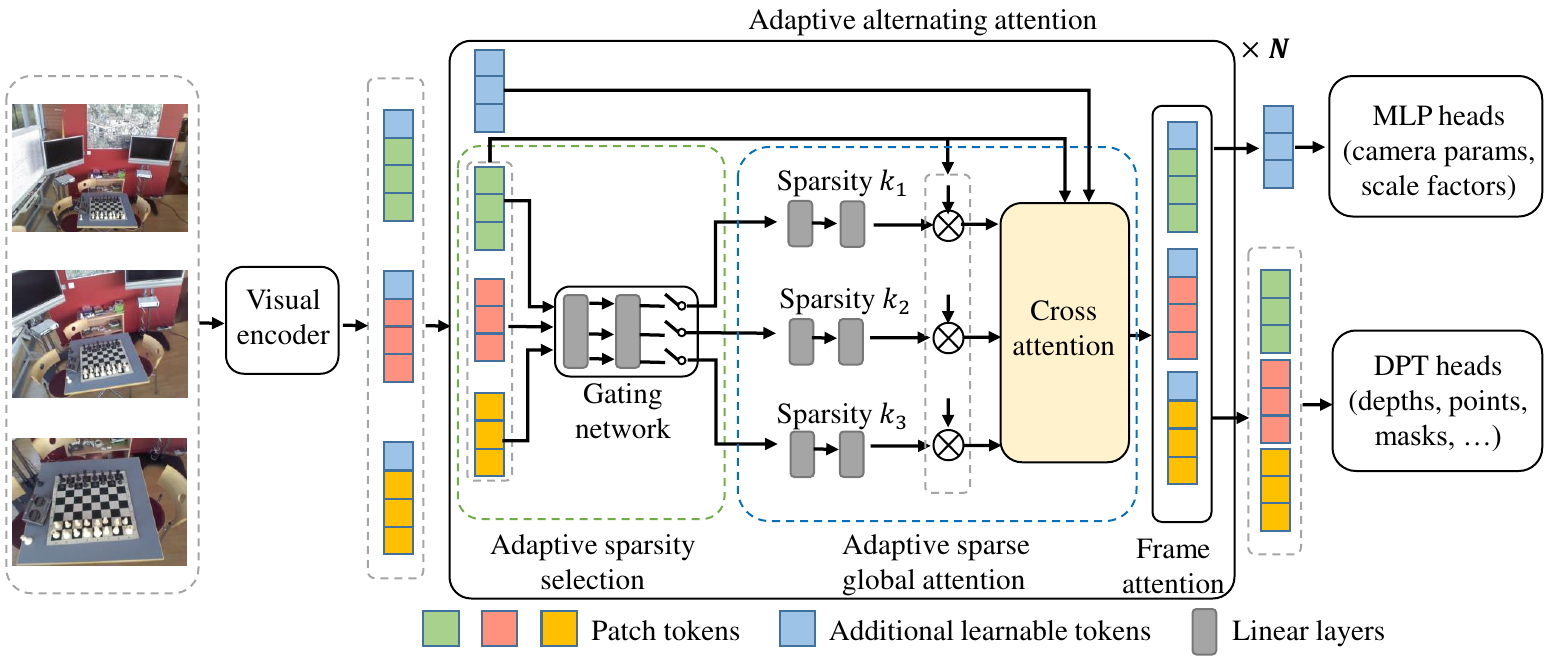}
    \caption{The overall framework of the proposed TurboVGGT.}
    \label{fig:framework}
\end{figure}

\subsection{Approach Overview}
An overview of the proposed TurboVGGT is depicted in Figure~\ref{fig:framework}.
The overall architecture of TurboVGGT consists of a visual encoder, $N$ adaptive alternating attention blocks, and multiple task-specific prediction heads.

\paragraph{Visual Encoder.}
Given a sequence of $L$ images $\{I_i\}_{i=1}^L$, we first employ a visual encoder (\eg, DINOv2~\cite{oquab2024dinov2}) to extract $M$ patch tokens from each image, denoted as $\{x_{i,j}\}_{j=1}^M$, where $x_{i,j} \in \mathbb{R}^D$ is a $D$-dimensional patch token.
In addition, following~\cite{wang2025vggt,keetha2026mapanything}, we also append additional learnable tokens, such as camera tokens and register tokens, to the sequence of patch tokens as the input to the subsequent modules.

\paragraph{Adaptive Alternating Attention Blocks.}
Next, we use $N$ blocks of the proposed adaptive alternating attention to process the input token sequence.
As shown in Figure~\ref{fig:framework}, adaptive alternating attention consists of adaptive sparsity selection, adaptive sparse global attention, and frame attention.
Specifically, in each block, we first aggregate the patch tokens for each frame and use a gating network to generate the gating score for each frame.
Based on gating scores, we route the patch tokens of each frame to the branch corresponding to a specific sparsity ratio $k_k$ and generate compressed representative tokens for each frame.
Then, we perform cross-attention between the compressed tokens and the dense tokens to capture global correspondences across frames.
After that, we perform frame attention for tokens from each frame to capture local geometric details within each frame.
This process is repeated for $N$ blocks.

\paragraph{Task-Specific Prediction Heads.}
Finally, we aggregate the output tokens from the adaptive alternating attention blocks and employ multiple task-specific prediction heads to decode the learned tokens into various 3D properties of each view.
As shown in Figure~\ref{fig:framework}, MLP heads~\cite{wang2025vggt,keetha2026mapanything} are used to predict camera parameters and metric scaling factors, while DPT heads~\cite{ranftl2021vision} are used to predict depth maps, point maps, confidence maps, \etc.

\subsection{Adaptive Alternating Attention}
\label{sec:method_adaptive_attention}

\begin{figure}[t]
    \centering
    \includegraphics[width=0.99\textwidth]{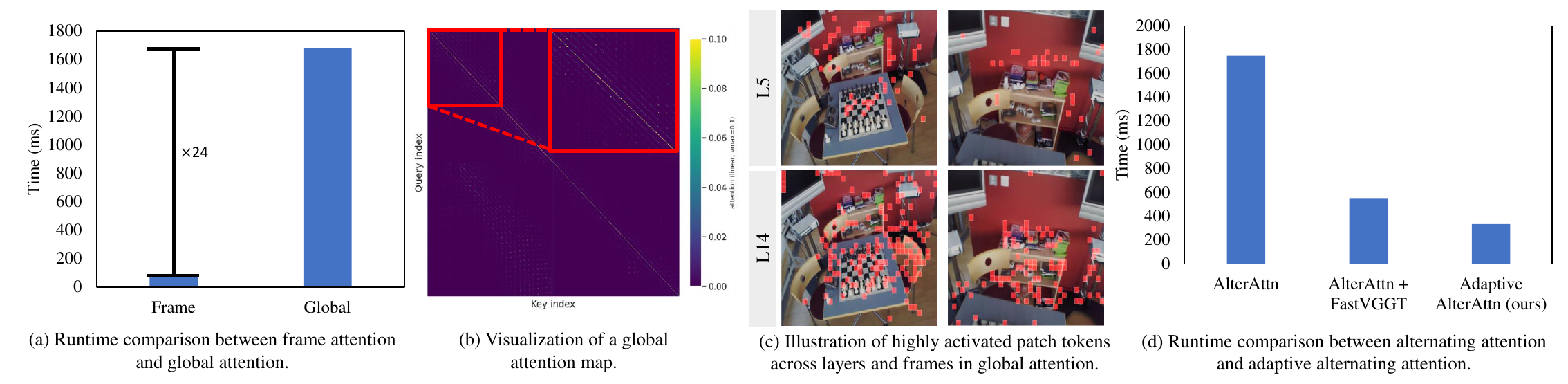}
    \caption{Motivation for our design.
    (a) The runtime comparison between a frame attention layer and a global attention layer during a single forward pass on 7-Scenes.
    (b) A visualization of an average global attention map in the alternating attention block.
    (c) An illustration of highly activated patch tokens (top 5\% of activations across all frames) across layers and frames in global attention.
    Note that visual geometry reconstruction requires not only feature matching and camera pose estimation, but also dense depth and pointmap prediction, so low-texture regions and globally informative background structures can also receive high responses.
    This example visualizes the top-activated tokens across different layers and frames, which may capture complementary geometric cues rather than consistently focusing on textured regions.
    (d) The runtime comparison between an alternating attention block and an adaptive alternating attention block during a single forward pass.
    Please refer to the appendix for enlarged visualizations.
    }
    \label{fig:motivation}
\end{figure}

\paragraph{Adaptive Sparsity Selection.}
State-of-the-art visual geometry transformers~\cite{wang2025vggt,keetha2026mapanything} typically employ alternating attention blocks consisting of frame-wise self-attention and cross-frame global full attention.
This design allows transformers to learn local geometric details within each frame and to learn global correspondences across frames in an alternating manner.
Although existing visual geometry transformers have achieved promising performance, they often suffer from long inference time and out-of-memory issues when processing long sequences.
One of the main reasons is that global full attention involves dense global token interactions, which have a quadratic computational complexity with respect to the number of input tokens~\cite{shen2026fastvggt,wang2025flashvggt,sun2025avggt,wang2025faster}.
As shown in Figure~\ref{fig:motivation}(a), a global attention layer consumes 24 times more runtime than a frame attention layer on the 7-Scenes dataset~\cite{shotton2013scene}.
Consequently, as the number of input frames increases, the runtime of the global attention layer increases significantly compared to the frame attention layer.
Nevertheless, as pointed out in prior work~\cite{wang2025faster,shen2026fastvggt}, the attention map of each global attention layer is typically sparse (see Figure~\ref{fig:motivation}(b)), where only a small portion of patch tokens are highly activated for learning global relationships across frames while the majority of patch tokens have low attention activations.
This suggests that it is not necessary to perform dense global full attention for all patch tokens across frames.
However, unlike prior works that use the same sparsity ratio for all frames in all global attention layers, we further analyze the highly activated patch tokens across layers and frames in global attention.
As shown in Figure~\ref{fig:motivation}(c), we find that the distribution of highly activated patch tokens can vary significantly across layers and frames.
This indicates that the same sparsity ratio for all frames in all global attention layers may not be optimal.
In light of this, we propose to adaptively select the sparsity level for each layer and frame rather than using a fixed sparsity ratio.
Specifically, as shown in Figure~\ref{fig:framework}, in each adaptive alternating attention block, we first aggregate the patch tokens for each frame.
Then, we use a gating network consisting of an MLP to generate the gating score for each frame and route each frame to the branch corresponding to a specific sparsity ratio $k_k=\{k_1, k_2, \dots, k_n\}$.
This is formulated as follows:
\begin{equation}
    k_k = F_s(F_g(F_{a}(\{x_{i,j}\}_{j=1}^M))),
\end{equation}
where $F_{a}(\cdot)$ is a function for generating frame-level representations, such as average pooling, $F_g(\cdot)$ is the gating network that generates the gating score for each frame, and $F_s(\cdot)$ is the softmax function that classifies each frame into one of the branches corresponding to a specific sparsity ratio $k_k$.

\paragraph{Adaptive Sparse Global Attention.}
Traditional multi-view 3D reconstruction methods~\cite{schonberger2016structure,schonberger2016pixelwise} typically capture global correspondences across frames by keypoint detection and matching rather than dense pixel-wise matching.
This suggests that global dependencies across frames can be achieved by focusing on structurally informative regions, such as keypoints and edges.
Moreover, as shown in Figure~\ref{fig:motivation}(c), highly activated patch tokens in global attention are sparsely distributed across frames and different frames may have different informative regions for learning global relationships.
This further indicates that learning representative tokens for each frame as structurally informative regions can not only reduce computations of redundant tokens but also facilitate global geometry modeling across frames.
In light of this, we propose to learn compressed representative tokens for each frame and employ cross-attention between the compressed tokens and the dense tokens to capture global correspondences across frames.
Specifically, in each branch corresponding to a specific sparsity ratio $k_k$, we employ an MLP $F_w(\cdot)$ to generate a weight matrix $W_k$ for tokens of each frame, which is defined as:
\begin{equation}
    W_i = F_w(x_{i}) + B_k,
\end{equation}
where $x_{i} {\in} \mathbb{R}^{M \times D}$ are the patch tokens of frame $i$, $W_i {\in} \mathbb{R}^{M \times M_{k_k}}$ is the weight matrix, $M_{k_k}{=}\lfloor M (1-k_k) \rfloor$ is the number of compressed representative tokens for the sparsity ratio $k_k$, and $B_k {\in} \mathbb{R}^{M \times M_{k_k}}$ is a learnable bias.
Then, we generate the compressed representative tokens for each frame by aggregating the patch tokens of each frame using the weight matrix $W_i$ as follows:
\begin{equation}
    x^c_i = W_i^T x_{i},
\end{equation}
where $x^c_i {\in} \mathbb{R}^{M_{k_k} \times D}$ are the compressed representative tokens for frame $i$.
Next, we concatenate the compressed representative tokens from all frames and additional learnable tokens (such as camera tokens and register tokens if used) as $x^c$.
Here, if a reference frame is used, tokens of the reference frame are also concatenated.
We then perform cross-attention between $x^c$ and all dense tokens $x'$ to capture global correspondences across frames, which is defined as follows:
\begin{equation}
    x'' = \text{CrossAttn}(x'W^Q, x^cW^K, x^cW^V),
\end{equation}
where $x''$ are the output dense tokens, $W^Q$, $W^K$, and $W^V$ are the projection matrices for query, key, and value, respectively.
In this way, we can capture global correspondences across frames while significantly reducing computational cost of global attention by using a small number of compressed representative tokens for each frame.

\paragraph{Remarks.}
TurboVGGT is not a straightforward combination of existing components, but a geometry-specific design for efficient visual geometry transformers.
While TurboVGGT shares high-level concepts with prior token compression and routing methods, its core mechanisms are not direct instantiations of prior methods, like Perceiver~\cite{jaegle2021perceiver}/Q-Former~\cite{li2023blip} or standard MoE. 
First, our gated router is not a standard MoE router that selects among functionally distinct experts.
Instead, it dynamically allocates token budgets across both frames and layers within global geometry attention.
Second, our representative tokens are input-conditioned, frame-specific weighted aggregations of dense patch tokens, rather than fixed latent queries shared across inputs as in Perceiver/Q-Former.
Dense tokens remain as queries to preserve dense geometric predictions, while only cross-frame keys and values are compressed.
Third, our proposed mechanisms are specifically coupled with the alternating attention architecture of VGGT, where dense frame attention preserves local geometry and adaptive compressed global attention captures cross-view correspondences.
Our experiments show that the straightforward combinations of conventional methods are insufficient for VGGT acceleration.

\subsection{Model Optimization}
Our TurboVGGT is trained in an end-to-end manner using a combination of losses for different prediction heads and a regularization term for the adaptive sparsity selection.
The overall loss is:
\begin{equation}
    \mathcal{L} = \mathcal{L}_{\text{recon}} + \lambda \mathcal{L}_{\text{reg}},
\end{equation}
where $\mathcal{L}_{\text{recon}}$ is a reconstruction loss, $\mathcal{L}_{\text{reg}}$ is a sparsity regularization loss, and $\lambda$ is a weight for the regularization term.
Specifically, for the reconstruction loss $\mathcal{L}_{\text{recon}}$, we follow prior work~\cite{wang2025vggt,wang2025pi,keetha2026mapanything} and use a sum of losses for different predictions, such as camera loss, depth loss, and pointmap loss, to supervise the learning of various 3D geometric properties.
Please refer to~\cite{wang2025vggt,wang2025pi,keetha2026mapanything} for the detailed definition of the reconstruction loss.
For the sparsity regularization loss $\mathcal{L}_{\text{reg}}$, it is used to encourage the model to select larger sparsity levels for different layers and frames (thereby reducing the computational cost).
By default, we use $\mathcal{L}_{\text{reg}} = \sum_{n=1}^N \sum_{i=1}^L (1-k_k^{n,i})$, where $k_k^{n,i}$ is the sparsity ratio selected for frame $i$ in the $n$-th adaptive alternating attention block.
Additionally, we can add an entropy term to the sparsity regularization loss to push each frame in each layer to route decisively to a specific branch, which can be defined as $\sum_{n=1}^{N}(-\frac{1}{L}\sum_{i=1}^{L} p^{n,i} \log p^{n,i})$, where $p^{n,i}$ is the probability of selecting a specific branch for frame $i$ in the $n$th block.

\section{Experiments}
\label{sec:experiments}

\begin{table}[t]
	\small
	\centering
    \caption{Evaluation of point cloud reconstruction on 7-Scenes, N-RGBD, and ScanNet.
    Inference timings for TurboVGGT, VGGT, FastVGGT, SparseVGGT, and Fast3R are measured on the same single data-center GPU system with approximately 140 GB of high-bandwidth memory under identical environmental settings and timing protocol.
    For timing, we perform multiple measured runs, with CUDA synchronization before and after each measurement, and report the average end-to-end inference time.
    *AVGGT results are taken from the original paper as the source code is not publicly available.
    }
	\tabcolsep=0.15cm
    \resizebox{0.99\textwidth}{!}{
	\begin{tabular}{c|ccccccc|ccccccc}
		\specialrule{1pt}{0pt}{2pt}   
		\specialrule{0.3pt}{0pt}{0pt} 
		\multirow{3}{*}{Method} & \multicolumn{7}{c|}{7-Scenes - Stride 3} & \multicolumn{7}{c}{7-Scenes - Stride 10} \\ \cline{2-15} 
		& \multicolumn{2}{c}{Acc~$\downarrow$} & \multicolumn{2}{c}{Comp~$\downarrow$} & \multicolumn{2}{c}{NC~$\uparrow$} & \multirow{2}{*}{Time $\downarrow$} & \multicolumn{2}{c}{Acc~$\downarrow$} & \multicolumn{2}{c}{Comp~$\downarrow$} & \multicolumn{2}{c}{NC~$\uparrow$} & \multirow{2}{*}{Time $\downarrow$} \\ \cline{2-7} \cline{9-14}
		& Mean & Med. & Mean & Med. & Mean & Med. &  & Mean & Med. & Mean & Med. & Mean & Med. &  \\ \hline
            Fast3R~\cite{yang2025fast3r} & 0.045 & {0.027} & {0.047} & \cellcolor{tabfirst}{0.010} & 0.616 & 0.627 & 21.2s & 0.040 & {0.021} & 0.056 & \cellcolor{tabsecond}{0.013} & \cellcolor{tabsecond}{0.639} & {0.657} & \cellcolor{tabsecond}{2.3s} \\
            VGGT~\cite{wang2025vggt} & \cellcolor{tabthird}{0.019} & \cellcolor{tabsecond}{0.009} & \cellcolor{tabsecond}{0.027} & \cellcolor{tabfirst}{0.010} & \cellcolor{tabthird}{0.622} & \cellcolor{tabthird}{0.699} & 38.1s & \cellcolor{tabthird}{0.019} & \cellcolor{tabfirst}{0.008} & \cellcolor{tabsecond}{0.027} & \cellcolor{tabfirst}{0.010} & 0.628 & \cellcolor{tabthird}{0.710} & 4.5s \\ 
            SparseVGGT~\cite{wang2025faster} & 0.021 & \cellcolor{tabthird}{0.010} & \cellcolor{tabthird}{0.029} & \cellcolor{tabfirst}{0.010} & 0.621 & 0.693 & \cellcolor{tabthird}{16.2s} & 0.020 & \cellcolor{tabthird}{0.010} & \cellcolor{tabthird}{0.028} & \cellcolor{tabfirst}{0.010} & 0.629 & 0.707 & \cellcolor{tabthird}{2.6s}  \\
            AVGGT~\cite{sun2025avggt}* & 0.054 & {0.031} & 0.131 & \cellcolor{tabsecond}{0.078} & 0.528 & 0.541 & {20.6s} & - & - & - & - & - & - & -  \\
            FastVGGT~\cite{shen2026fastvggt} &  \cellcolor{tabsecond}{0.018} & \cellcolor{tabfirst}{0.008} & \cellcolor{tabfirst}{0.026} & \cellcolor{tabfirst}{0.010} & \cellcolor{tabsecond}{0.627} & \cellcolor{tabsecond}{0.707} & \cellcolor{tabsecond}{14.2s} & \cellcolor{tabsecond}{0.018} & \cellcolor{tabfirst}{0.008} & \cellcolor{tabsecond}{0.027} & \cellcolor{tabfirst}{0.010} & \cellcolor{tabthird}{0.632} & \cellcolor{tabsecond}{0.717} & {2.7s} \\
            TurboVGGT (ours) & \cellcolor{tabfirst}{0.016} & \cellcolor{tabsecond}{0.009} & \cellcolor{tabfirst}{0.026} & \cellcolor{tabfirst}{0.010} & \cellcolor{tabfirst}{0.639} & \cellcolor{tabfirst}{0.714} & \cellcolor{tabfirst}{9.6s} & \cellcolor{tabfirst}{0.016} & \cellcolor{tabsecond}{0.009} & \cellcolor{tabfirst}{0.025} & \cellcolor{tabfirst}{0.010} & \cellcolor{tabfirst}{0.650} & \cellcolor{tabfirst}{0.730} & \cellcolor{tabfirst}{2.0s} \\
       \specialrule{1pt}{0pt}{2pt}  
        \specialrule{0.3pt}{0pt}{0pt}
        \multirow{3}{*}{Method} & \multicolumn{7}{c|}{N-RGBD - Stride 3} & \multicolumn{7}{c}{N-RGBD - Stride 10} \\ \cline{2-15} 
        & \multicolumn{2}{c}{Acc~$\downarrow$} & \multicolumn{2}{c}{Comp~$\downarrow$} & \multicolumn{2}{c}{NC~$\uparrow$} & \multirow{2}{*}{Time $\downarrow$} & \multicolumn{2}{c}{Acc~$\downarrow$} & \multicolumn{2}{c}{Comp~$\downarrow$} & \multicolumn{2}{c}{NC~$\uparrow$} & \multirow{2}{*}{Time $\downarrow$} \\ \cline{2-7} \cline{9-14}
        & Mean & Med. & Mean & Med. & Mean & Med. &  & Mean & Med. & Mean & Med. & Mean & Med. &  \\ \hline
            Fast3R~\cite{yang2025fast3r} & 0.074 & 0.036 & 0.024 & \cellcolor{tabsecond}{0.011} & \cellcolor{tabsecond}{0.658} & 0.682 & 37.0s & 0.061 & 0.028 & 0.031 & {0.013} & \cellcolor{tabsecond}{0.669} & 0.712 & \cellcolor{tabsecond}{3.9s} \\
            VGGT~\cite{wang2025vggt} & \cellcolor{tabsecond}{0.028} & \cellcolor{tabsecond}{0.018} & \cellcolor{tabfirst}{0.018} & \cellcolor{tabfirst}{0.010} & \cellcolor{tabthird}{0.657} & \cellcolor{tabsecond}{0.779} & 65.3s & \cellcolor{tabfirst}{0.016} & \cellcolor{tabfirst}{0.010} & \cellcolor{tabfirst}{0.016} & \cellcolor{tabfirst}{0.009} & \cellcolor{tabsecond}{0.669} & \cellcolor{tabfirst}{0.791} & 7.2s \\
            SparseVGGT~\cite{wang2025faster} & 0.045 & 0.029 & 0.024 & \cellcolor{tabsecond}{0.011} & 0.647 & 0.755 & \cellcolor{tabsecond}{26.2s} & 0.041 & 0.027 & 0.027 & \cellcolor{tabthird}{0.012} & 0.652 & \cellcolor{tabthird}{0.763} & {4.1s} \\                          
            FastVGGT~\cite{shen2026fastvggt} & \cellcolor{tabthird}{0.031} & \cellcolor{tabthird}{0.020} & \cellcolor{tabsecond}{0.019} & \cellcolor{tabfirst}{0.010} & \cellcolor{tabfirst}{0.662} & \cellcolor{tabfirst}{0.788} & \cellcolor{tabthird}{30.2s} & \cellcolor{tabsecond}{0.017} & \cellcolor{tabsecond}{0.011} & \cellcolor{tabsecond}{0.017} & \cellcolor{tabfirst}{0.009} & \cellcolor{tabfirst}{0.671} & \cellcolor{tabfirst}{0.791} & \cellcolor{tabthird}{4.0s} \\
            TurboVGGT (ours) & \cellcolor{tabfirst}{0.025} & \cellcolor{tabfirst}{0.016} & \cellcolor{tabthird}{0.022} & \cellcolor{tabsecond}{0.011} & \cellcolor{tabthird}{0.657} & \cellcolor{tabthird}{0.773} & \cellcolor{tabfirst}{14.7s} & \cellcolor{tabthird}{0.021} & \cellcolor{tabthird}{0.014} & \cellcolor{tabthird}{0.021} & \cellcolor{tabsecond}{0.011} & \cellcolor{tabthird}{0.664} & \cellcolor{tabsecond}{0.783} & \cellcolor{tabfirst}{2.9s} \\
       \specialrule{1pt}{0pt}{2pt}  
        \specialrule{0.3pt}{0pt}{0pt}
        \multirow{2}{*}{Method} & \multicolumn{4}{c|}{ScanNet - 500 Frames} & \multicolumn{4}{c|}{ScanNet - 300 Frames} & \multicolumn{4}{c|}{ScanNet - 100 Frames} & \multicolumn{2}{c}{Average}\\ \cline{2-15} 
        & \multicolumn{2}{c}{CD~$\downarrow$} & \multicolumn{2}{c|}{Time $\downarrow$} & \multicolumn{2}{c}{CD~$\downarrow$} & \multicolumn{2}{c|}{Time $\downarrow$} & \multicolumn{2}{c}{CD~$\downarrow$} & \multicolumn{2}{c|}{Time $\downarrow$} & {CD~$\downarrow$} & {Time $\downarrow$}\\ \hline
            Fast3R~\cite{yang2025fast3r} & \multicolumn{2}{c}{0.701} & \multicolumn{2}{c|}{62.5s} & \multicolumn{2}{c}{0.711} & \multicolumn{2}{c|}{17.2s} & \multicolumn{2}{c}{0.723} & \multicolumn{2}{c|}{\cellcolor{tabsecond}{2.5s}} & {0.712} & {27.4s} \\
            VGGT~\cite{wang2025vggt} & \multicolumn{2}{c}{0.464} & \multicolumn{2}{c|}{90.1s} & \multicolumn{2}{c}{0.454} & \multicolumn{2}{c|}{33.6s} & \multicolumn{2}{c}{\cellcolor{tabthird}{0.438}} & \multicolumn{2}{c|}{4.9s} & \cellcolor{tabthird}{0.452} & {42.9s} \\
            SparseVGGT~\cite{wang2025faster} & \multicolumn{2}{c}{\cellcolor{tabthird}{0.459}} & \multicolumn{2}{c|}{\cellcolor{tabthird}{34.2s}} & \multicolumn{2}{c}{\cellcolor{tabthird}{0.453}} & \multicolumn{2}{c|}{\cellcolor{tabthird}{14.2s}} & \multicolumn{2}{c}{0.446} & \multicolumn{2}{c|}{\cellcolor{tabthird}{2.6s}} & {0.453} & \cellcolor{tabthird}{17.0s} \\
            FastVGGT~\cite{shen2026fastvggt} & \multicolumn{2}{c}{\cellcolor{tabsecond}{0.453}} & \multicolumn{2}{c|}{\cellcolor{tabsecond}{28.4s}} & \multicolumn{2}{c}{\cellcolor{tabsecond}{0.447}} & \multicolumn{2}{c|}{\cellcolor{tabsecond}{12.3s}} & \multicolumn{2}{c}{\cellcolor{tabsecond}{0.425}} & \multicolumn{2}{c|}{{2.8s}} & \cellcolor{tabsecond}{0.442} & \cellcolor{tabsecond}{14.5s} \\
            TurboVGGT (ours) & \multicolumn{2}{c}{\cellcolor{tabfirst}{0.416}} & \multicolumn{2}{c|}{\cellcolor{tabfirst}{20.3s}} & \multicolumn{2}{c}{\cellcolor{tabfirst}{0.413}} & \multicolumn{2}{c|}{\cellcolor{tabfirst}{9.5s}} & \multicolumn{2}{c}{\cellcolor{tabfirst}{0.400}} & \multicolumn{2}{c|}{\cellcolor{tabfirst}{2.2s}} & \cellcolor{tabfirst}{0.410} & \cellcolor{tabfirst}{10.7s} \\
       \hline
	\end{tabular}
    }    
    \label{tab:point_cloud_reconstruction}
\end{table}

\subsection{Datasets and Experimental Setup}

\paragraph{Datasets.}
We train our TurboVGGT on 13 high-quality datasets~\cite{keetha2026mapanything} spanning diverse indoor, outdoor, and in-the-wild scenes, including BlendedMVS~\cite{yao2020blendedmvs}, Mapillary Planet-Scale Depth~\cite{antequera2020mapillary}, ScanNet++ v2~\cite{yeshwanth2023scannet++}, Spring~\cite{mehl2023spring}, TartanAirV2-WB~\cite{wang2020tartanair,zhang2025ufm}, UnrealStereo4K~\cite{tosi2021smd}, Aria Synthetic Environments~\cite{avetisyan2024scenescript}, DL3DV-10K~\cite{ling2024dl3dv}, Dynamic Replica~\cite{karaev2023dynamicstereo}, MegaDepth~\cite{li2018megadepth}, MVS-Synth~\cite{huang2018deepmvs}, ParallelDomain-4D~\cite{van2024generative}, and SAIL-VOS 3D~\cite{hu2021sail}.
We randomly resize and crop image sequences from these training datasets to various aspect ratios with a maximum resolution of 518 pixels.
We evaluate our TurboVGGT on several challenging benchmark datasets, including 7-Scenes~\cite{shotton2013scene}, N-RGBD~\cite{azinovic2022neural}, ScanNet-50~\cite{dai2017scannet}, RealEstate10K~\cite{zhou2018stereo}, and Sintel~\cite{butler2012naturalistic}.

\paragraph{Evaluation Metrics.}
Following prior work~\cite{shen2026fastvggt}, we evaluate our TurboVGGT on three tasks, including point cloud reconstruction, camera pose estimation, and depth estimation.
For point cloud reconstruction, we report Accuracy distance (Acc, $\downarrow$), Completeness (Comp, $\downarrow$), and Normal Consistency (NC, $\uparrow$) on 7-Scenes and N-RGBD, and Chamfer Distance (CD, $\downarrow$) on ScanNet.
For camera pose estimation, we report Relative Rotation Accuracy (RRA@30, $\uparrow$), Relative Translation Accuracy (RTA@30, $\uparrow$), and the Area Under the accuracy-threshold Curve for the minimum of RRA and RTA (AUC@30, $\uparrow$) on 7-Scenes, N-RGBD, and RealEstate10K.
For depth estimation, we report Absolute Relative Error (AbsRel, $\downarrow$) and $\delta{<}1.25$ ratio ($\uparrow$) on 7-Scenes, N-RGBD, and Sintel.
In addition, we also report inference time (seconds, $\downarrow$).

\paragraph{Implementation Details.}
We implement our TurboVGGT with Python and PyTorch.
By default, we use VGGT~\cite{wang2025vggt} as the backbone and also adapt our approach to other backbones, \ie, MapAnything~\cite{keetha2026mapanything} (denoted as TurboVGGT-M) and $\pi^3$~\cite{wang2025pi} (denoted as TurboVGGT-$\pi$), to validate the generalizability of our approach.
We employ DINOv2~\cite{oquab2024dinov2} as the visual encoder and build the adaptive alternating attention blocks as described in Sec.~\ref{sec:method_adaptive_attention}.
For the adaptive sparsity selection, we use three branches and set $k_k = \{\frac{3}{4}, \frac{8}{9}, \frac{15}{16}\}$, \ie, use three sparsity ratios of $75\%$, $89\%$, and $94\%$ for the three branches, respectively, because these sparsity levels correspond to the token budgets induced by 2$\times$, 3$\times$, and 4$\times$ spatial compression, respectively.
These levels cover conservative, moderate, and aggressive computation budgets.
In addition, discrete levels produce regular token shapes and computation patterns, and a small fixed set of sparsity levels simplifies optimization and prevents the gating network from selecting extreme token budgets.
The number of the adaptive alternating attention blocks is determined by the backbone~\cite{wang2025vggt,wang2025pi,keetha2026mapanything}.
We construct the task-specific prediction heads following~\cite{wang2025vggt,wang2025pi,keetha2026mapanything}.
To facilitate model convergence, we initialize our TurboVGGT with pretrained weights from~\cite{wang2025vggt,wang2025pi,keetha2026mapanything}.
We train our TurboVGGT on 8 GPUs for 10 epochs using AdamW optimizer with a cosine annealing learning rate and a 1-epoch warm-up schedule.
The maximum learning rate is set to $5{\times}10^{-6}$ and decayed to $5{\times}10^{-8}$.
For FastVGGT, we use the merging ratio of 90\%, and for SparseVGGT, we use the sparsity ratio of 75\%.

\begin{table}[t]
	\small
	\centering
    \caption{Evaluation of camera pose estimation on 7-Scenes, N-RGBD, and RealEstate10K.
    }
	\tabcolsep=0.15cm
    \resizebox{0.99\textwidth}{!}{
	\begin{tabular}{c|cccc|cccc|ccc}
		\specialrule{1pt}{0pt}{2pt}   
		\specialrule{0.3pt}{0pt}{0pt} 
		\multirow{2}{*}{Method} & \multicolumn{4}{c|}{7-Scenes (dense)} & \multicolumn{4}{c|}{N-RGBD (dense)} & \multicolumn{3}{c}{RealEstate10K} \\\cline{2-12}
		& RRA@30$\uparrow$ & RTA@30$\uparrow$ & AUC@30$\uparrow$ & Time $\downarrow$ & RRA@30$\uparrow$ & RTA@30$\uparrow$ & AUC@30$\uparrow$ & Time $\downarrow$ & RRA@30$\uparrow$ & RTA@30$\uparrow$ & AUC@30$\uparrow$ \\
        \hline
        Fast3R~\cite{yang2025fast3r} & -& -& -&- & -& -& -&- & 99.05 & 81.86 & 61.68 \\
        FLARE~\cite{zhang2025flare} & -& -& -&- & -& -& -&- & 99.69 & 95.23 & 80.01 \\
        VGGT~\cite{wang2025vggt} & \cellcolor{tabfirst}{100.00} & \cellcolor{tabsecond}{96.58} & \cellcolor{tabthird}{77.76} & {38.1s} & \cellcolor{tabfirst}{100.00} & \cellcolor{tabthird}{99.47} & \cellcolor{tabthird}{91.59} & 65.3s & \cellcolor{tabfirst}{99.97} & \cellcolor{tabfirst}{96.22} & \cellcolor{tabsecond}{85.32} \\
        SparseVGGT~\cite{wang2025faster} &\cellcolor{tabfirst}{100.00} &93.40 &70.07 & \cellcolor{tabthird}{16.2s} & \cellcolor{tabsecond}{97.20}& 97.03& 83.16& \cellcolor{tabsecond}{26.2s} & 99.61 & 91.35 & 76.65 \\
        AVGGT~\cite{sun2025avggt} &\cellcolor{tabfirst}{100.00} & \cellcolor{tabthird}{96.48} & \cellcolor{tabsecond}{78.29} & 20.6s & -& -& -&- & \cellcolor{tabsecond}{99.93} & \cellcolor{tabsecond}{95.62} & \cellcolor{tabfirst}{85.45} \\
        Speed3R~\cite{ren2026speed3r} & - & - & - & - & - & - & - & - & - & - & 74.81 \\
        FlashVGGT~\cite{wang2025flashvggt} & -& -& -& - & -& -& -&- & \cellcolor{tabthird}{99.92} & \cellcolor{tabthird}{95.61} & \cellcolor{tabthird}{85.30} \\
        FastVGGT~\cite{shen2026fastvggt} & \cellcolor{tabsecond}{99.99}& 96.29& 76.90& \cellcolor{tabsecond}{14.2s} & \cellcolor{tabfirst}{100.00}& \cellcolor{tabsecond}{99.65}& \cellcolor{tabsecond}{92.47}& \cellcolor{tabthird}{30.2s} & \cellcolor{tabthird}{99.92} & 94.76 & 84.37 \\
        TurboVGGT (ours) & \cellcolor{tabfirst}{100.00} & \cellcolor{tabfirst}{96.83} & \cellcolor{tabfirst}{81.87} & \cellcolor{tabfirst}{9.6s}& \cellcolor{tabfirst}{100.00} & \cellcolor{tabfirst}{99.71} & \cellcolor{tabfirst}{93.28} & \cellcolor{tabfirst}{14.7s} & \cellcolor{tabsecond}{99.93} & 94.69 & 84.31 \\
        \hline
	\end{tabular}
    }    
    \label{tab:camera_pose_estimation}
\end{table}
\begin{table}[t]
	\small
	\centering
    \caption{Evaluation of depth estimation on 7-Scenes, N-RGBD, and Sintel.}
	\tabcolsep=0.15cm
    \resizebox{0.75\textwidth}{!}{
	\begin{tabular}{c|ccc|ccc|cc}
		\specialrule{1pt}{0pt}{2pt}   
		\specialrule{0.3pt}{0pt}{0pt} 
		\multirow{2}{*}{Method} & \multicolumn{3}{c|}{7-Scenes (dense)} & \multicolumn{3}{c|}{N-RGBD (dense)} & \multicolumn{2}{c}{Sintel} \\\cline{2-9}
		& AbsRel$\downarrow$ & $\delta{<}1.25$$\uparrow$ & Time $\downarrow$ & AbsRel$\downarrow$ & $\delta{<}1.25$$\uparrow$ & Time $\downarrow$ & AbsRel$\downarrow$ & $\delta{<}1.25$$\uparrow$ \\
        \hline
        Fast3R~\cite{yang2025fast3r} & - & - & - & - & - & - & 0.544 & 0.509 \\
        FLARE~\cite{zhang2025flare} & - & - & - & - & - & - & 0.606 & 0.402 \\
        VGGT~\cite{wang2025vggt} & \cellcolor{tabfirst}{0.264} & \cellcolor{tabsecond}{0.958} & {38.1s} & \cellcolor{tabfirst}{0.013} & \cellcolor{tabsecond}{0.993} & {65.3s} & \cellcolor{tabsecond}{0.335} & \cellcolor{tabsecond}{0.599} \\
        SparseVGGT~\cite{wang2025faster} & \cellcolor{tabthird}{0.393} & \cellcolor{tabthird}{0.956} & \cellcolor{tabthird}{16.2s} & \cellcolor{tabsecond}{0.015} & \cellcolor{tabthird}{0.990} & \cellcolor{tabsecond}{26.2s} & - & - \\
        FlashVGGT~\cite{wang2025flashvggt} & - & - & - & - & - & - & 0.346 & \cellcolor{tabthird}{0.586} \\
        FastVGGT~\cite{shen2026fastvggt} & 0.394 & 0.953 & \cellcolor{tabsecond}{14.2s} & \cellcolor{tabfirst}{0.013} & \cellcolor{tabthird}{0.990} & \cellcolor{tabthird}{30.2s} & \cellcolor{tabthird}{0.337} & 0.582 \\                
        TurboVGGT (ours) & \cellcolor{tabsecond}{0.296}  & \cellcolor{tabfirst}{0.980}  & \cellcolor{tabfirst}{9.6s}  & \cellcolor{tabfirst}{0.013} & \cellcolor{tabfirst}{0.994} & \cellcolor{tabfirst}{14.7s} & \cellcolor{tabfirst}{0.287}  & \cellcolor{tabfirst}{0.650}  \\
        \hline
	\end{tabular}
    }    
    \label{tab:depth_estimation}
\end{table}

\subsection{Comparison with State-of-the-Art Methods}

\begin{figure}[t]
    \centering
    \includegraphics[width=0.99\linewidth]{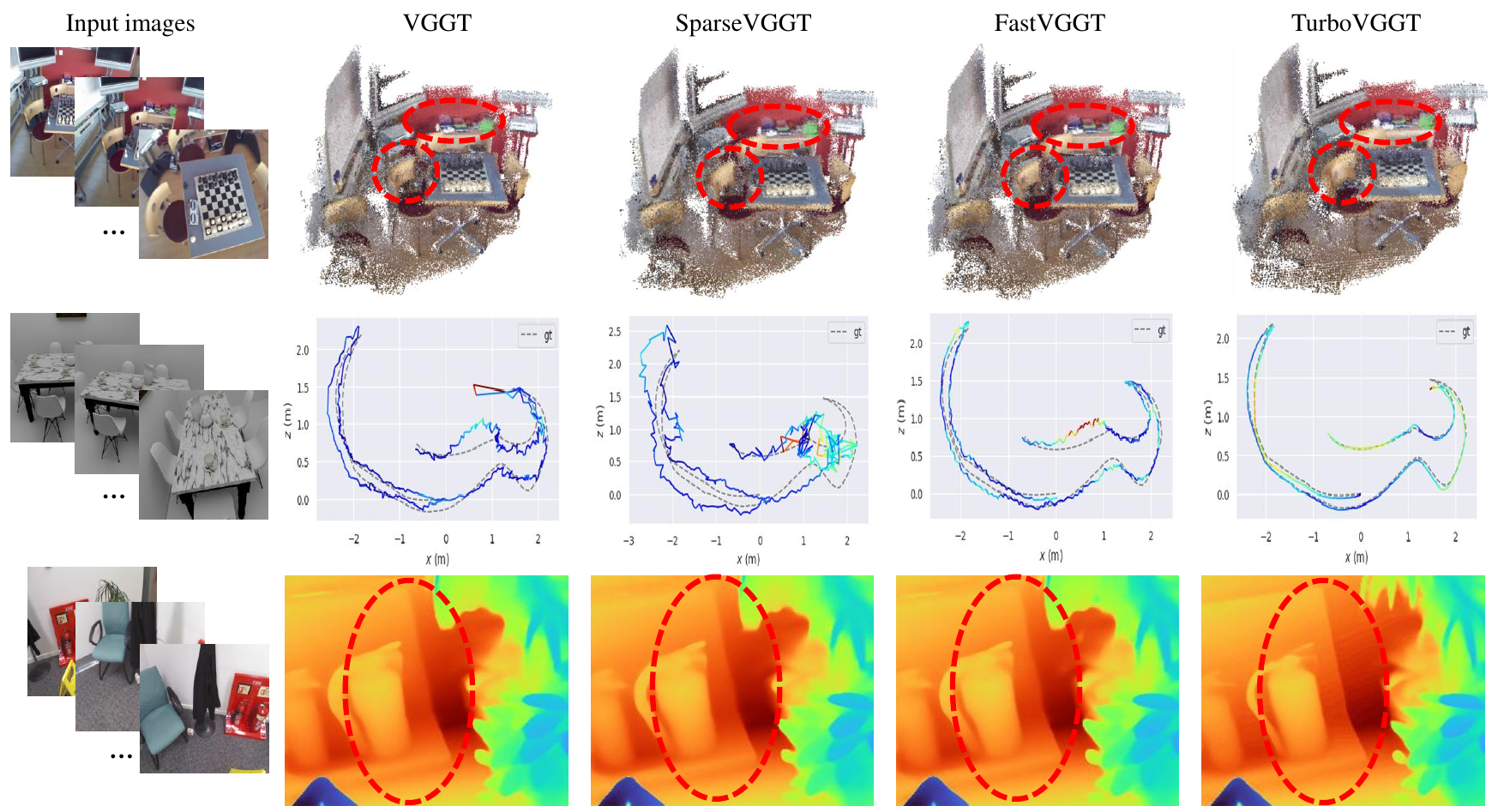}
    \caption{Qualitative comparison of point cloud reconstruction, camera pose estimation, and depth estimation (7-Scenes and N-RGBD with stride 3).
    See appendix for more qualitative results.}

    \label{fig:sota_qualitative}
\end{figure}

\paragraph{Point Cloud Reconstruction.}
Table~\ref{tab:point_cloud_reconstruction} shows results of point cloud reconstruction on 7-Scenes, N-RGBD, and ScanNet.
Overall, our TurboVGGT achieves faster inference speed than state-of-the-art methods, while maintaining comparable or even better performance.
Specifically, on 7-Scenes, our TurboVGGT outperforms state-of-the-art methods, such as FastVGGT~\cite{shen2026fastvggt}, SparseVGGT~\cite{wang2025faster}, and VGGT~\cite{wang2025vggt}, in most metrics.
Interestingly, TurboVGGT achieves overall better reconstruction quality than VGGT~\cite{wang2025vggt}, while being 2-4$\times$ faster.
This can be attributed to the adaptive sparse global attention under the guidance of adaptive sparsity selection, which effectively captures global context while reducing redundant computations.
On N-RGBD, our TurboVGGT also achieves competitive reconstruction performance while achieving faster inference than prior methods.
Although the reconstruction quality of TurboVGGT is slightly inferior to FastVGGT in some metrics, TurboVGGT is 1.4$\times$ faster than FastVGGT in the Stride 10 setting and 2.1$\times$ faster in the Stride 3 setting.
On ScanNet, TurboVGGT achieves the best reconstruction quality while being faster than prior methods.
Figure~\ref{fig:sota_qualitative} (first row) shows qualitative comparison of point cloud reconstruction.
We can see that our TurboVGGT produces more complete point clouds compared with state-of-the-art methods.

\paragraph{Camera Pose Estimation.}
We report results of camera pose estimation on 7-Scenes (dense, stride=3), N-RGBD (dense, stride=3), and RealEstate10K (sparse) in Table~\ref{tab:camera_pose_estimation}.
On 7-Scenes and N-RGBD, our TurboVGGT achieves the best camera pose estimation performance and faster inference speed compared with state-of-the-art methods, such as FastVGGT~\cite{shen2026fastvggt}, SparseVGGT~\cite{wang2025faster}, and VGGT~\cite{wang2025vggt}.
In addition, we also evaluate TurboVGGT on RealEstate10K, which contains sparse image sequences.
Our TurboVGGT still achieves competitive performance in this sparse setting.
These results show the efficiency and effectiveness of our TurboVGGT in camera pose estimation across different settings.
Figure~\ref{fig:sota_qualitative} (second row) shows that the camera pose trajectory produced by our TurboVGGT is closer to the ground-truth trajectory than state-of-the-art competitors.

\paragraph{Depth Estimation.}
Table~\ref{tab:depth_estimation} shows results of depth estimation on 7-Scenes (dense), N-RGBD (dense), and Sintel (single).
On 7-Scenes, our TurboVGGT achieves the second-best AbsRel and the best $\delta{<}1.25$, while being faster than prior methods.
On N-RGBD, our TurboVGGT achieves the best results in all metrics.
Additionally, we evaluate TurboVGGT for monocular depth estimation on Sintel.
Our TurboVGGT still achieves competitive performance in this monocular depth estimation setting, demonstrating the generalizability of our approach across settings.
Figure~\ref{fig:sota_qualitative} (third row) shows that our TurboVGGT produces higher-quality depth compared with state-of-the-art methods.

\begin{table}[t]
	\small
	\centering
    \caption{Evaluation of our approach with various backbones on 7-Scenes (stride 3).}
	\tabcolsep=0.15cm
    \resizebox{0.99\textwidth}{!}{
	\begin{tabular}{c|cc|cc|c|c}
		\specialrule{1pt}{0pt}{2pt}   
		\specialrule{0.3pt}{0pt}{0pt} 
		\multirow{2}{*}{Method} 
        & \multicolumn{2}{c|}{Point Cloud Estimation} 
        & \multicolumn{2}{c|}{Camera Pose Estimation} 
        & \multicolumn{1}{c|}{Depth Estimation} 
        & \multirow{2}{*}{Time $\downarrow$}\\\cline{2-6}
		& Acc$\downarrow$ 
        & NC$\uparrow$ 
        & RRA@30$\uparrow$ 
        & RTA@30$\uparrow$ 
        & $\delta{<}1.25$$\uparrow$ 
        & \\
        \hline
        
        VGGT~\cite{wang2025vggt} 
        & 0.019 & 0.622 
        & 100.00 & 96.58 
        & 0.958 & 38.1s \\
        
        VGGT~\cite{wang2025vggt} (Fine-tuned) 
        & 0.016 & 0.638 
        & 100.00 & 96.57 
        & 0.960 & 38.1s \\
        
        TurboVGGT (Ours) 
        & 0.016 & 0.639 
        & 100.00 & 96.83 
        & 0.980 & \textbf{9.6s} (4.0$\times$) \\
        \hline
        
        $\pi^3$~\cite{wang2025pi} 
        & 0.013 & 0.585 
        & 100.00 & 97.47 
        & 0.986 & 30.4s \\
        
        $\pi^3$~\cite{wang2025pi} (Fine-tuned) 
        & 0.015 & 0.586 
        & 100.00 & 97.50 
        & 0.984 & 30.4s \\
        
        TurboVGGT-$\pi$ (Ours) 
        & 0.014 & 0.585 
        & 100.00 & 97.11 
        & 0.983 & \textbf{6.2s} (4.9$\times$) \\
        \hline
        
        MapAnything~\cite{keetha2026mapanything} 
        & 0.018 & 0.579 
        & 100.00 & 92.45 
        & 0.975 & 15.3s \\
        
        MapAnything~\cite{keetha2026mapanything} (Fine-tuned) 
        & 0.021 & 0.569 
        & 99.99 & 90.60 
        & 0.974 & 15.3s \\
        
        TurboVGGT-M (Ours) 
        & 0.018 & 0.577 
        & 100.00 & 92.86 
        & 0.975 & \textbf{4.0s} (3.8$\times$) \\
        \hline
	\end{tabular}
    }    
    \label{tab:backbone_evaluation}
\end{table}

\subsection{Adaptation to Different Backbones}
To validate the generalizability of the proposed adaptive alternating attention, we adapt our TurboVGGT to more visual geometry transformer backbones, including $\pi^3$~\cite{wang2025pi} and MapAnything~\cite{keetha2026mapanything}.
Results are shown in Table~\ref{tab:backbone_evaluation}.
We can see that our approach consistently improves the inference speed of these backbones while maintaining comparable or even better performance on point cloud reconstruction, camera pose estimation, and depth estimation.
This demonstrates the effectiveness and generalizability of our adaptive alternating attention in accelerating visual geometry transformers across different backbones.

\subsection{Ablation Study}
\paragraph{Component Effectiveness.}
We present the ablation study of our TurboVGGT on 7-Scenes in Table~\ref{tab:ablation_study}.
For the adaptive sparsity selection, we report two variants of TurboVGGT, \ie, V1 replaces the adaptive sparsity selection with fixed selection that evenly assigns frames to the three branches, and V2 removes the adaptive sparsity selection and uses only one branch with a fixed sparsity ratio.
From Table~\ref{tab:ablation_study}, we can see that both V1 and V2 achieve worse performance than the full TurboVGGT.
This shows the importance of the adaptive sparsity selection in routing frames to different branches for better capturing global context and reducing redundant computations.
For the adaptive sparse global attention, we also report two variants of TurboVGGT, \ie, V3 replaces weight matrix based compressed representative token learning with grid based token selection, and V4 removes the adaptive sparse global attention and uses global attention for merged tokens and upsampling to restore the original token number.
From Table~\ref{tab:ablation_study}, we can see that without using weight matrix based compressed representative token learning, the reconstruction quality of V3 is worse than the full TurboVGGT.
When removing the adaptive sparse global attention, V4 achieves significantly worse reconstruction quality than the full TurboVGGT.
These results demonstrate the importance of the adaptive sparse global attention in learning representative tokens and performing global attention on these tokens for better capturing global relationships and reducing redundant computations.

\paragraph{Branch Sparsity Ratio Selection.}
In Table~\ref{tab:sparsity_ratio}, we report additional ablation study on the branch sparsity ratio selection, including one, two, three, and four sparsity ratios.
Using three sparsity ratios achieves the best speed-accuracy trade-off performance;
When using one sparsity ratio of 75\%, the reconstruction quality is comparable to the default three-sparsity-ratio setting, but the inference speed is significantly slower;
When using two high sparsity ratios can reduce inference time but slightly degrade accuracy;
When using four sparsity ratios, the inference time is even faster, but the reconstruction quality is significantly degraded.

\newcommand{\cmark}{\ding{51}}
\newcommand{\xmark}{\ding{55}}

\begin{table}[t]
    \small
    \centering  
    \caption{Ablation study of the proposed TurboVGGT on 7-Scenes (stride 3).}  
    \tabcolsep=0.15cm
    \resizebox{0.72\textwidth}{!}{
    \begin{tabular}{c|cccc|ccc}
        \specialrule{1pt}{0pt}{2pt}   
		\specialrule{0.3pt}{0pt}{0pt}      
        \multirow{2}{*}{Method} & \multicolumn{2}{c}{Ada. Selection} & \multicolumn{2}{c|}{Ada. Global Attn} & \multirow{2}{*}{Acc$\downarrow$} & \multirow{2}{*}{AUC@30$\uparrow$} & \multirow{2}{*}{AbsRel$\downarrow$} \\
        & Adaptive & Multi-branch & Weight matrix & Cross-attn\\
        \midrule               
        V1 & \xmark   & \cmark        & \cmark         & \cmark     & \textbf{0.016} & {81.34} & 0.302 \\
        V2 & \xmark   & \xmark        & \cmark         & \cmark     & \textbf{0.016} & 80.88 & 0.300 \\
        V3 & \cmark   & \cmark        & \xmark         & \cmark     & {0.017} & 79.80 & {0.299} \\
        V4 & \cmark   & \cmark        & \xmark         & \xmark     & 0.047 & 65.73 & 0.311 \\
        Full & \cmark   & \cmark        & \cmark         & \cmark     & \textbf{0.016} & \textbf{81.87} & \textbf{0.296} \\        
        \bottomrule
    \end{tabular}
    }
\label{tab:ablation_study}
\end{table}

\begin{table}[t]
	\small
	\centering
    \caption{Ablation study of different sparsity ratios on 7-Scenes (stride 3).}
	\tabcolsep=0.15cm
    \resizebox{0.99\textwidth}{!}{
	\begin{tabular}{c|cc|cc|c|c}
		\specialrule{1pt}{0pt}{2pt}   
		\specialrule{0.3pt}{0pt}{0pt} 
		\multirow{2}{*}{Variants} 
        & \multicolumn{2}{c|}{Point Cloud Estimation} 
        & \multicolumn{2}{c|}{Camera Pose Estimation} 
        & \multicolumn{1}{c|}{Depth Estimation} 
        & \multirow{2}{*}{Time $\downarrow$}\\\cline{2-6}
		& Acc$\downarrow$ 
        & NC$\uparrow$ 
        & RRA$\uparrow$ 
        & RTA$\uparrow$ 
        & $\delta{<}1.25$$\uparrow$ 
        & \\
        \hline
        Three sparsity ratios (75\%, 89\%, 94\%) (default)
        & 0.016 & \textbf{0.639} & \textbf{100.00} & \textbf{96.83} & \textbf{0.980} & 9.6s \\
        One sparsity ratio (75\%) 
        & \textbf{0.015} & 0.625 & \textbf{100.00} & 96.81 & \textbf{0.980} & 16.1s \\
        One sparsity ratio (89\%) 
        & 0.016 & 0.621 & \textbf{100.00} & 96.59 & 0.979 & 10.0s \\        
        Two sparsity ratios (89\%, 94\%) 
        & 0.018 & 0.619 & \textbf{100.00} & 96.32 & 0.979 & 8.9s \\
        Four sparsity ratios (75\%, 89\%, 94\%, 96\%) 
        & 0.019 & 0.617 & \textbf{100.00} & 95.48 & 0.978 & \textbf{8.2s} \\
        \hline
	\end{tabular}
    }    
    \label{tab:sparsity_ratio}
\end{table}

\section{Conclusion}
In this work, we propose TurboVGGT for fast multi-view 3D reconstruction in a feed-forward manner.
The core idea of TurboVGGT is to construct an efficient visual geometry transformer with adaptive alternating attention blocks, which consist of adaptive sparse global attention guided by adaptive sparsity selection for global geometry modeling and frame attention for local detail aggregation.
Extensive experiments on multiple 3D reconstruction benchmarks demonstrate that TurboVGGT achieves a better balance between reconstruction quality and computational efficiency, outperforming state-of-the-art methods.

{
 \small
 \bibliographystyle{plainnat}
  \bibliography{ref}
}


\appendix

\section{Technical appendices and supplementary material}

\subsection{Limitation and Future Work}
While the proposed TurboVGGT achieves superior performance on fast multi-view 3D reconstruction, there are still some limitations that can be addressed in future work.
First, TurboVGGT does not explicitly model temporal dynamics of input sequences, which may limit its performance on highly dynamic scenes.
Second, although TurboVGGT achieves a balance between reconstruction quality and computational efficiency, there is still room for improvement, especially for challenging scenes with complex geometry.
Third, the variable token counts in TurboVGGT are inconvenient to batch and parallelize.
For the associated training cost, TurboVGGT has lower peak-memory requirements but higher training compute cost (as shown in Table~\ref{tab:training_inference_efficiency}).
Compared with the fine-tuned VGGT, TurboVGGT reduces the observed peak allocated GPU memory from 100 GB/GPU to 70 GB/GPU, enabling a lower-memory training footprint.
This comes at additional optimization time cost because of the routing mechanism, adaptive sparse global attention, and gradient checkpointing.
However, at inference, a single scene uses hard argmax routing and tightly packs the resulting variable-length tokens, which provides the intended efficiency gain.
Thus, TurboVGGT primarily provides inference efficiency and reduced training memory, rather than faster training.
In future work, we aim to explore more effective ways to model temporal dynamics for 4D reconstruction and improve the model architecture design of TurboVGGT for better performance.
\begin{table}[h]
	\small
	\centering
    \caption{Comparison of training and inference efficiency on 7-Scenes (stride 3).}
	\tabcolsep=0.15cm
    \resizebox{0.8\textwidth}{!}{
	\begin{tabular}{c|cc|cc}
		\specialrule{1pt}{0pt}{2pt}
		\specialrule{0.3pt}{0pt}{0pt}
		\multirow{2}{*}{Method}
        & \multicolumn{2}{c|}{Training}
        & \multicolumn{2}{c}{Inference} \\ \cline{2-5}
		& Time
        & Peak GPU Memory
        & Time
        & Peak GPU Memory \\
        \hline
        Fine-tuned VGGT
        & \textbf{16.6 h}
        & 100 GB
        & 14.2 s
        & 25 GB \\

        TurboVGGT (Ours)
        & 22.7 h
        & \textbf{70 GB}
        & \textbf{9.6 s}
        & \textbf{23 GB} \\
        \hline
	\end{tabular}
    }
    \label{tab:training_inference_efficiency}
\end{table}

\subsection{Societal Impacts}
This work presents an efficient visual geometry transformer for multi-view 3D reconstruction, which can potentially benefit various applications such as virtual reality, robotics, and autonomous driving.
We do not discover direct negative societal impacts of our approach.
However, since our approach requires training with large-scale datasets and substantial computational resources, it may raise concerns regarding data privacy and environmental impact.
These societal impacts should be carefully considered when using our approach in real-world applications.

\subsection{Complexity Analysis of Adaptive Sparse Global Attention}
As discussed in the main paper, global full attention used in conventional visual geometry transformers~\cite{wang2025vggt} is computationally expensive and can lead to long inference time and out-of-memory issues when processing long sequences.
The computational complexity of global full attention is $\mathcal{O}(L^2 M^2)$, which is quadratic with respect to the number of input tokens, where $L$ is the number of input frames and $M$ is the number of patch tokens for each frame.
For our adaptive sparse global attention, the computational cost mainly comes from the cross-attention between the compressed representative tokens and the dense tokens, while the other operations have relatively less computational cost (especially for long sequences) which can be ignored for simplicity.
Specifically, the computational complexity of our adaptive sparse global attention is $\mathcal{O}(L^2 M^2 (1-k_k))$, where $k_k$ is the sparsity ratio.
Since we encourage our model to select a large sparsity ratio $k_k$ for each frame and have $(1-k_k){\ll}1$, \eg, $(1-k_k) = \{25\%, 11\%, 6\%\}$, our adaptive sparse global attention significantly reduces the computational complexity compared to global full attention, which is consistent with the significant runtime reduction observed in our experiments (see Section~\ref{sec:experiments}).
Although some existing works~\cite{shen2026fastvggt,wang2025flashvggt} also have a computational complexity similar to ours, they typically use the same sparsity ratio for all frames across layers and lack mechanisms to adaptively learn representative tokens that can effectively capture global relationships; in contrast, our method adaptively selects the sparsity level across layers and frames and achieves better reconstruction quality while maintaining fast inference speed.

\subsection{Memory Efficiency Analysis}
In Table~\ref{tab:suppl_memory}, we compare the peak GPU memory usage of our TurboVGGT, VGGT~\cite{wang2025vggt}, SparseVGGT~\cite{wang2025faster}, and FastVGGT~\cite{shen2026fastvggt} on 7-Scenes (dense setting).
We can see that our TurboVGGT not only achieves faster inference speed but also reduces the peak GPU memory usage compared with these state-of-the-art methods.
This result further shows the superiority of our TurboVGGT that can effectively reduce the computational cost and memory usage for multi-view 3D reconstruction.
\begin{table}[h]
    \centering 
    \caption{Memory efficiency analysis on 7-Scenes.}
    \tabcolsep=0.15cm
    \resizebox{0.95\textwidth}{!}{
    \begin{tabular}{c|cccc}
        \specialrule{1pt}{0pt}{2pt}   
		\specialrule{0.3pt}{0pt}{0pt}        
        Metrics                             & VGGT~\cite{wang2025vggt} & SparseVGGT~\cite{wang2025faster} & FastVGGT~\cite{shen2026fastvggt} & TurboVGGT (Ours) \\
        \hline        
        Peak inference memory$\downarrow$   & 25.24 GB   & 27.84 GB   & 31.18 GB   & \textbf{23.47 GB} \\
        Inference speed$\uparrow$           & 8.27 FPS  & 19.56 FPS & 22.16 FPS &  \textbf{33.01 FPS} \\
        \hline
    \end{tabular}
    }     
    \label{tab:suppl_memory}
\end{table}

\subsection{Evaluation of Video Depth Estimation}
In Table~\ref{tab:suppl_video_depth}, we further perform evaluation of video depth estimation on Sintel~\cite{butler2012naturalistic} and Bonn~\cite{palazzolo2019refusion}, following~\cite{elflein2026vgg}.
We can see that our TurboVGGT achieves the best performance on both datasets, outperforming state-of-the-art methods such as VGG-T$^3$~\cite{elflein2026vgg} and ZipMap~\cite{jin2026zipmap}.
This further demonstrates the superior performance of our TurboVGGT in video depth estimation, which is crucial for multi-view 3D reconstruction.
\begin{table}[h]
    \small
	\centering
    \caption{Evaluation of video depth estimation on Sintel and Bonn.
    Results of some methods are taken from~\cite{elflein2026vgg,jin2026zipmap}.
    }
	\tabcolsep=0.15cm
    \resizebox{0.99\textwidth}{!}{   
    \begin{tabular}{c|c|ccccccc}
    \specialrule{1pt}{0pt}{2pt}   
	\specialrule{0.3pt}{0pt}{0pt} 
    Datasets & Metrics & VGGT~\cite{wang2025vggt} & SparseVGGT~\cite{wang2025faster} & FastVGGT~\cite{shen2026fastvggt} & TTT3R~\cite{chen2026tttr} & ZipMap~\cite{jin2026zipmap} & VGG-T$^3$~\cite{elflein2026vgg} & TurboVGGT (Ours)\\
    \hline
    \multirow{2}{*}{Sintel} 
    & AbsRel$\downarrow$        & 0.300 & 0.304 & 0.307 & 0.469 & 0.248 & 0.345 & \textbf{0.212}  \\
    & $\delta{<}1.25$$\uparrow$ & 0.646 & 0.639 & 0.630 & 0.510 & 0.695 & 0.581 & \textbf{0.716} \\
    \hline
    \multirow{2}{*}{Bonn} 
    & AbsRel$\downarrow$        & 0.059 & 0.057 & 0.058 & 0.061 & 0.059 & 0.063 & \textbf{0.053} \\
    & $\delta{<}1.25$$\uparrow$ & 0.967 & 0.968 & 0.969 & 0.969 & 0.973 & 0.963 & \textbf{0.975} \\
    \hline
    \end{tabular}
    }
    \label{tab:suppl_video_depth}
\end{table}

\subsection{Impact of Sparsity Regularization Loss $\mathcal{L}_{reg}$}
In Table~\ref{tab:suppl_sparsity_reg_loss}, we analyze the impact of the sparsity regularization loss $\mathcal{L}_{reg}$ on our TurboVGGT.
When setting $\lambda_{reg}=0$, our TurboVGGT does not have an explicit regularization to encourage sparsity, resulting in both worse reconstruction quality and slower inference speed compared with our full model.
When setting $\lambda_{reg}=0.001$,  the inference speed is improved compared with $\lambda_{reg}=0$, but the reconstruction quality is still worse than our full model.
When setting $\lambda_{reg}=0.01$, our TurboVGGT achieves the best performance in terms of both reconstruction quality and inference speed.
\begin{table}[h]
    \centering 
    \caption{Impact of sparsity regularization loss $\mathcal{L}_{reg}$ (on 7-Scenes).}
    \tabcolsep=0.15cm
    \resizebox{0.9\textwidth}{!}{
    \begin{tabular}{c|cccc}
        \specialrule{1pt}{0pt}{2pt}   
		\specialrule{0.3pt}{0pt}{0pt}       
        Method         &  {Acc$\downarrow$}  &  {AUC@30$\uparrow$}  & AbsRel$\downarrow$ & {Time$\downarrow$} \\
        \hline               
        TurboVGGT w/o regularization loss ($\lambda_{reg}=0$) & \textbf{0.016} & 80.63 & 0.309 & 14.5s \\
        TurboVGGT ($\lambda_{reg}=0.001$)                              & 0.017 & 80.23 & 0.301 & 12.5s  \\
        TurboVGGT (default, $\lambda_{reg}=0.01$)                               & \textbf{0.016} & \textbf{81.87} & \textbf{0.296}& \textbf{9.6s} \\
        \hline
    \end{tabular}
    }     
    \label{tab:suppl_sparsity_reg_loss}
\end{table}

\subsection{Impact of Separating Additional Learnable Tokens}
In our design, we separate additional learnable tokens of each frame from dense patch tokens during adaptive sparsity selection and compressed representative token learning.
This allows our model to focus on capturing global relationships across frames without being distracted by additional global tokens.
In Table~\ref{tab:suppl_separate_token}, we report a variant of our TurboVGGT that does not separate the additional learnable tokens during adaptive sparsity selection and compressed representative token learning.
It can be seen that this variant performs slightly worse than our full model for multi-view 3D reconstruction.
\begin{table}[h]
    \centering 
    \caption{Impact of separating additional learnable tokens (on 7-Scenes).}
    \tabcolsep=0.15cm
    \resizebox{0.95\textwidth}{!}{
    \begin{tabular}{c|cccccc}
        \specialrule{1pt}{0pt}{2pt}   
		\specialrule{0.3pt}{0pt}{0pt}       
        Method         &  {Acc$\downarrow$}  &  {Comp$\downarrow$} & {RRA@30$\uparrow$} & {RTA@30$\uparrow$} & {AbsRel$\downarrow$} & $\delta{<}1.25$$\uparrow$ \\
        \hline       
        TurboVGGT      & \textbf{0.016} & 0.026 & \textbf{100.00} & \textbf{96.83} & \textbf{0.296} & \textbf{0.980} \\
        TurboVGGT w/o separating additional tokens       & 0.017 & \textbf{0.024} & \textbf{100.00} & 96.29 & 0.314 & 0.979 \\
        \hline
    \end{tabular}
    }     
    \label{tab:suppl_separate_token}
\end{table}

\subsection{Difference from Q-Former and Perceiver}
Q-Former~\cite{li2023blip} and Perceiver~\cite{jaegle2021perceiver} employ a compact set of latent queries and cross-attention to perform attention over the input in a low-rank latent space, which shares some merits with our adaptive sparse global attention design.
However, our approach significantly differs from them in motivation and design.
Q-Former and Perceiver use a set of learnable latents for all inputs to transform the inputs into a compact latent space, which may not be optimal for multi-view 3D reconstruction where the importance of different frames can vary significantly;
in contrast, our approach adaptively selects the sparsity level across layers and frames and learns compressed representative tokens for each frame.
More importantly, our method is specifically designed for multi-view 3D reconstruction, which requires modeling global correspondences across frames while preserving local geometric details within each frame;
in contrast, Q-Former and Perceiver are designed for general language and computer vision tasks and do not explicitly consider the unique challenges of multi-view 3D reconstruction.
As shown Table~\ref{tab:suppl_qformer}, we report a variant of VGGT that uses a set of learnable query latents.
We can see that this variant achieves significantly worse reconstruction quality compared with our TurboVGGT.
This result further demonstrates the importance of our adaptive sparse global attention in TurboVGGT for achieving superior performance, while using existing design such as learnable query latents is not sufficient to capture representative tokens across frames for multi-view 3D reconstruction.
\begin{table}[h]
    \centering 
    \caption{Comparison between the proposed design and a variant of VGGT that uses learnable query latents (point cloud reconstruction on 7-Scenes (stride 3)).}
    \tabcolsep=0.15cm
    \resizebox{0.95\textwidth}{!}{
    \begin{tabular}{c|cccccc}
        \specialrule{1pt}{0pt}{2pt}   
		\specialrule{0.3pt}{0pt}{0pt}       
        Method         &  {Acc$\downarrow$}  &  {Comp$\downarrow$} & {RRA@30$\uparrow$} & {RTA@30$\uparrow$} & {AbsRel$\downarrow$} & $\delta{<}1.25$$\uparrow$ \\
        \hline      
        TurboVGGT                         & \textbf{0.016} & \textbf{0.026} & \textbf{100.00} & \textbf{96.83} & \textbf{0.296} & \textbf{0.980} \\
        VGGT + learnable query latents   & 0.109 & 0.041 & 91.05 & 78.24 & 0.362 & 0.975 \\
        \hline
    \end{tabular}
    }     
    \label{tab:suppl_qformer}
\end{table}

\subsection{Discrete Sparsity Ratio vs. Continuous Sparsity Ratio}
In our default setting, we use three discrete sparsity ratios for the branches rather than a continuous sparsity formulation because we consider that although a continuous sparsity formulation can be more flexible, it may lead to irregular token shapes and computation patterns, which can increase optimization difficulty.
In Table~\ref{tab:discrete_continuous_sparsity}, we additionally evaluate a continuous sparsity variant.
Our results show that a continuous sparsity formulation achieves slightly worse performance than the default discrete sparsity formulation.
\begin{table}[h]
	\small
	\centering
    \caption{Ablation study of discrete and continuous sparsity formulations on 7-Scenes (stride 3).}
	\tabcolsep=0.15cm
    \resizebox{0.95\textwidth}{!}{
	\begin{tabular}{c|cc|cc|c}
		\specialrule{1pt}{0pt}{2pt}   
		\specialrule{0.3pt}{0pt}{0pt} 
		\multirow{2}{*}{Method} 
        & \multicolumn{2}{c|}{Point Cloud Estimation} 
        & \multicolumn{2}{c|}{Camera Pose Estimation} 
        & \multicolumn{1}{c}{Depth Estimation} \\\cline{2-6}
		& Acc$\downarrow$ 
        & NC$\uparrow$ 
        & RRA$\uparrow$ 
        & RTA$\uparrow$ 
        & $\delta{<}1.25$$\uparrow$ \\
        \hline
        Three discrete sparsity ratios 
        & \textbf{0.016} & \textbf{0.639} 
        & \textbf{100.00} & \textbf{96.83} 
        & \textbf{0.980} \\
        
        Continuous sparsity formulation 
        & 0.017 & 0.618 
        & \textbf{100.00} & 96.76 
        & 0.978 \\
        \hline
	\end{tabular}
    }    
    \label{tab:discrete_continuous_sparsity}
\end{table}

\subsection{Evaluation on Camera Tracking Accuracy}
In this work, we mainly report AUC, RRA, and RTA for camera pose estimation, following the common evaluation protocol of recent visual geometry transformers.
These metrics support flexible, potentially sparse or unordered image inputs.
In Table~\ref{tab:camera_tracking}, we additionally evaluate camera tracking accuracy with ATE$\downarrow$, RPE-rot$\downarrow$, and RPE-trans$\downarrow$ metrics on 7-Scenes, Sintel, and Bonn datasets.
Our results demonstrate that TurboVGGT is faster than VGGT and achieves superior camera tracking accuracy compared with VGGT on 7-Scenes and Sintel, while maintaining competitive performance on Bonn.
\begin{table}[h]
	\small
	\centering
    \caption{Evaluation of camera tracking accuracy on 7-Scenes, Sintel, and Bonn.}
	\tabcolsep=0.15cm
    \resizebox{0.7\textwidth}{!}{
	\begin{tabular}{c|ccc|c}
		\specialrule{1pt}{0pt}{2pt}   
		\specialrule{0.3pt}{0pt}{0pt} 
		\multirow{2}{*}{Method} 
        & \multicolumn{3}{c|}{7-Scenes - Stride 3} 
        & \multirow{2}{*}{Inference Time$\downarrow$} \\ \cline{2-4}
		& ATE$\downarrow$ 
        & RPE-rot$\downarrow$ 
        & RPE-trans$\downarrow$ 
        & \\
        \hline
        VGGT 
        & 0.087 & 0.502 & \textbf{0.022} & 38.1s \\
        TurboVGGT(Ours) 
        & \textbf{0.082} & \textbf{0.498} & \textbf{0.022} & \textbf{9.6s} \\

        \specialrule{1pt}{0pt}{2pt}   
		\specialrule{0.3pt}{0pt}{0pt} 
		\multirow{2}{*}{Method} 
        & \multicolumn{3}{c|}{Sintel} 
        & \multirow{2}{*}{Inference Time$\downarrow$} \\ \cline{2-4}
		& ATE$\downarrow$ 
        & RPE-rot$\downarrow$ 
        & RPE-trans$\downarrow$ 
        & \\
        \hline
        VGGT 
        & 0.274 & 0.473 & 0.094 & 1.8s \\
        TurboVGGT(Ours) 
        & \textbf{0.210} & \textbf{0.464} & \textbf{0.059} & \textbf{1.5s} \\

        \specialrule{1pt}{0pt}{2pt}   
		\specialrule{0.3pt}{0pt}{0pt} 
		\multirow{2}{*}{Method} 
        & \multicolumn{3}{c|}{Bonn} 
        & \multirow{2}{*}{Inference Time$\downarrow$} \\ \cline{2-4}
		& ATE$\downarrow$ 
        & RPE-rot$\downarrow$ 
        & RPE-trans$\downarrow$ 
        & \\
        \hline
        VGGT 
        & \textbf{0.013} & \textbf{0.294} & \textbf{0.007} & 7.5s \\
        TurboVGGT(Ours) 
        & \textbf{0.013} & 0.304 & 0.009 & \textbf{3.5s} \\
        \hline
	\end{tabular}
    }    
    \label{tab:camera_tracking}
\end{table}

\subsection{Comparison with FastVGGT and SparseVGGT under Different Sparsity Ratios}
Table~\ref{tab:sota_higher_sparse} presents the comparison with FastVGGT and SparseVGGT under different merging/sparsity rates on 7-Scenes (stride 3).
The results show that TurboVGGT achieves better speed-accuracy trade-off than FastVGGT and SparseVGGT under all merging/sparsity ratios.
It is worth noting that FastVGGT has token merging, salient token selection, reference token selection, token unmerging, and some other operations, so increasing the merging ratio from 90\% to a higher value (e.g., 94\%) does not necessarily lead to faster inference speed.
For SparseVGGT, increasing the sparsity ratio from 75\% to a higher value (e.g., 90\%) lead to significantly worse reconstruction quality and accuracy though the inference speed is improved.
\begin{table}[h]
	\small
	\centering
    \caption{Comparison with FastVGGT and SparseVGGT under different sparsity ratios on 7-Scenes (stride 3).}
	\tabcolsep=0.15cm
    \resizebox{0.9\textwidth}{!}{
	\begin{tabular}{c|cc|cc|c|c}
		\specialrule{1pt}{0pt}{2pt}   
		\specialrule{0.3pt}{0pt}{0pt} 
		\multirow{2}{*}{Method} 
        & \multicolumn{2}{c|}{Point Cloud Estimation} 
        & \multicolumn{2}{c|}{Camera Pose Estimation} 
        & \multicolumn{1}{c|}{Depth Estimation} 
        & \multirow{2}{*}{Time $\downarrow$}\\\cline{2-6}
		& Acc$\downarrow$ 
        & NC$\uparrow$ 
        & RRA$\uparrow$ 
        & RTA$\uparrow$ 
        & $\delta{<}1.25$$\uparrow$ 
        & \\
        \hline
        TurboVGGT (Ours) 
        & \textbf{0.016} 
        & \textbf{0.639} 
        & \textbf{100.00} 
        & \textbf{96.83} 
        & \textbf{0.980} 
        & \textbf{9.6s} \\
        
        FastVGGT (90\%) 
        & 0.018 
        & 0.627 
        & 99.99 
        & 96.29 
        & 0.953 
        & 14.2s \\
        
        FastVGGT (94\%) 
        & 0.018 
        & 0.626 
        & \textbf{100.00} 
        & 96.29 
        & 0.953 
        & 14.1s \\
        
        SparseVGGT (75\%) 
        & 0.021 
        & 0.621 
        & \textbf{100.00} 
        & 93.40 
        & 0.956 
        & 16.2s \\
        
        SparseVGGT (90\%) 
        & 0.034 
        & 0.615 
        & 97.31 
        & 92.79 
        & 0.948 
        & 10.6s \\
        \hline
	\end{tabular}
    }    
    \label{tab:sota_higher_sparse}
\end{table}

\subsection{Comparison with Fine-tuned FastVGGT and SparseVGGT}
FastVGGT and SparseVGGT are proposed as post-hoc acceleration methods, so our main evaluation faithfully follows their official implementations by applying them directly to the vanilla VGGT.
End-to-end fine-tuning of them is not defined in the original methods and would require additional method-specific training choices.
However, fine-tuning them may also improve their performance.
In Table~\ref{tab:finetune_sota}, we further report the performance of fine-tuned FastVGGT and SparseVGGT using the same training setup as TurboVGGT.
The results show that TurboVGGT achieves better speed-accuracy trade-off than FastVGGT (Fine-tuned) and SparseVGGT (Fine-tuned), demonstrating the effectiveness of our approach.
\begin{table}[h]
	\small
	\centering
    \caption{Comparison with FastVGGT and SparseVGGT on 7-Scenes (stride 3).}
	\tabcolsep=0.12cm
    \resizebox{0.98\textwidth}{!}{
	\begin{tabular}{c|cc|cc|c|c|cc}
		\specialrule{1pt}{0pt}{2pt}   
		\specialrule{0.3pt}{0pt}{0pt} 
		\multirow{2}{*}{Method} 
        & \multicolumn{2}{c|}{Point Cloud Estimation} 
        & \multicolumn{2}{c|}{Camera Pose Estimation} 
        & \multicolumn{1}{c|}{Depth Estimation} 
        & \multirow{2}{*}{Inference Time$\downarrow$} 
        & \multicolumn{2}{c}{Peak GPU Memory$\downarrow$} \\\cline{2-6}\cline{8-9}
		& Acc$\downarrow$ 
        & NC$\uparrow$ 
        & RRA$\uparrow$ 
        & RTA$\uparrow$ 
        & $\delta{<}1.25$$\uparrow$ 
        & 
        & Inference 
        & Training \\
        \hline
        TurboVGGT (Ours) 
        & 0.016 
        & \textbf{0.639} 
        & \textbf{100.00} 
        & \textbf{96.83} 
        & \textbf{0.980} 
        & \textbf{9.6s} 
        & \textbf{23GB} 
        & \textbf{70GB} \\
        
        FastVGGT (Official) 
        & 0.018 
        & 0.627 
        & 99.99 
        & 96.29 
        & 0.953 
        & 14.2s 
        & 31GB 
        & - \\
        
        FastVGGT (Fine-tuned) 
        & 0.017 
        & 0.620 
        & \textbf{100.00} 
        & 96.80 
        & 0.979 
        & 14.2s 
        & 31GB 
        & 121GB \\
        
        SparseVGGT (Official) 
        & 0.021 
        & 0.621 
        & \textbf{100.00} 
        & 93.40 
        & 0.956 
        & 16.2s 
        & 28GB 
        & - \\
        
        SparseVGGT (Fine-tuned, reimplemented sparse attention) 
        & \textbf{0.015} 
        & 0.623 
        & \textbf{100.00} 
        & 96.46 
        & 0.979 
        & 170.9s 
        & 50GB 
        & 102GB \\
        \hline
	\end{tabular}
    }    
    \label{tab:finetune_sota}
\end{table}

\subsection{Comparison with Fine-tuned FastVGGT Under Higher Merging Rate}
We further fine-tune FastVGGT with a higher merging rate of 94\% and reported the results in Table~\ref{tab:ft_fastvggt_higher_merge}.
The results show that TurboVGGT still achieves better speed-accuracy trade-off than FastVGGT under the higher merging rate.
It is worth noting that FastVGGT has token merging, salient token selection, reference token selection, token unmerging, and some other operations, so increasing the merging ratio from 90\% to a higher value (e.g., 94\%) does not necessarily lead to better speed-accuracy trade-off.
\begin{table}[h]
	\small
	\centering
    \caption{Comparison with fine-tuned FastVGGT under higher merging rate on 7-Scenes (stride 3).}
	\tabcolsep=0.12cm
    \resizebox{0.98\textwidth}{!}{
	\begin{tabular}{c|cc|cc|c|c|cc}
		\specialrule{1pt}{0pt}{2pt}   
		\specialrule{0.3pt}{0pt}{0pt} 
		\multirow{2}{*}{Method} 
        & \multicolumn{2}{c|}{Point Cloud Estimation} 
        & \multicolumn{2}{c|}{Camera Pose Estimation} 
        & \multicolumn{1}{c|}{Depth Estimation} 
        & \multirow{2}{*}{Inference Time$\downarrow$} 
        & \multicolumn{2}{c}{Peak GPU Memory$\downarrow$} \\\cline{2-6}\cline{8-9}
		& Acc$\downarrow$ 
        & NC$\uparrow$ 
        & RRA$\uparrow$ 
        & RTA$\uparrow$ 
        & $\delta{<}1.25$$\uparrow$ 
        & 
        & Inference 
        & Training \\
        \hline
        TurboVGGT (Ours) 
        & \textbf{0.016} 
        & \textbf{0.639} 
        & \textbf{100.00} 
        & \textbf{96.83} 
        & \textbf{0.980} 
        & \textbf{9.6s} 
        & \textbf{23GB} 
        & \textbf{70GB} \\
        
        FastVGGT (Fine-tuned, 90\%) 
        & 0.017 
        & 0.620 
        & \textbf{100.00} 
        & 96.80 
        & 0.979 
        & 14.2s 
        & 31GB 
        & 121GB \\
        
        FastVGGT (Fine-tuned, 94\%) 
        & 0.017 
        & 0.619 
        & \textbf{100.00} 
        & 96.80 
        & 0.979 
        & 14.1s 
        & 31GB 
        & 121GB \\
        \hline
	\end{tabular}
    }    
    \label{tab:ft_fastvggt_higher_merge}
\end{table}

\subsection{Comparison with Co-Me}
Co-Me~\cite{chen2026co} uses a distilled confidence predictor to identify and merge low-confidence tokens for efficient visual geometry transformers.
In contrast, TurboVGGT aims to adaptively allocate sparsity across frames and layers and learns representative tokens for cross-view geometry aggregation.
Thus, Co-Me focuses on confidence-guided token merging without modifying the pretrained model, whereas TurboVGGT learns an adaptive compressed-attention mechanism during training.
We further compare TurboVGGT with Co-Me in Table~\ref{tab:suppl_comparison_co_me}.
Inference time is measured with prediction heads included.
Following the official implementations provided by the authors, we run the experiments for Co-Me, including the CVPR26 version and the latest accelerated version.
These results show that TurboVGGT achieves better speed-accuracy trade-off than Co-Me.
\begin{table}[h]
	\small
	\centering
    \caption{Comparison with Co-Me on 7-Scenes (stride 3).}
	\tabcolsep=0.15cm
    \resizebox{0.9\textwidth}{!}{
	\begin{tabular}{c|cc|cc|c|c}
		\specialrule{1pt}{0pt}{2pt}   
		\specialrule{0.3pt}{0pt}{0pt} 
		\multirow{2}{*}{Method} 
        & \multicolumn{2}{c|}{Point Cloud Estimation} 
        & \multicolumn{2}{c|}{Camera Pose Estimation} 
        & \multicolumn{1}{c|}{Depth Estimation} 
        & \multirow{2}{*}{Time $\downarrow$}\\\cline{2-6}
		& Acc$\downarrow$ 
        & NC$\uparrow$ 
        & RRA$\uparrow$ 
        & RTA$\uparrow$ 
        & $\delta{<}1.25$$\uparrow$ 
        & \\
        \hline
        TurboVGGT (Ours) 
        & \textbf{0.016} 
        & \textbf{0.639} 
        & \textbf{100.00} 
        & \textbf{96.83} 
        & \textbf{0.980} 
        & \textbf{9.6s} \\
        
        Co-Me (CVPR26 version) 
        & 0.026 
        & 0.565 
        & \textbf{100.00} 
        & 95.19 
        & 0.950 
        & 23.8s \\
        
        Co-Me (latest accelerated version) 
        & 0.027 
        & 0.566 
        & \textbf{100.00} 
        & 94.60 
        & 0.945 
        & 16.4s \\
        \hline
	\end{tabular}
    }    
    \label{tab:suppl_comparison_co_me}
\end{table}

\subsection{Long Sequence Reconstruction Quality}
In Table~\ref{tab:long_sequence_7scenes}, we further report the reconstruction quality at long sequences of 500 frames and 1000 frames on the 7-Scenes dataset.
The results show that TurboVGGT achieves competitive reconstruction quality compared to the baselines while achieving faster inference speed, demonstrating its effectiveness in long-sequence scenarios.
\begin{table}[h]
	\small
	\centering
    \caption{Evaluation on long-sequence 7-Scenes with 1000 and 500 frames.}
	\tabcolsep=0.15cm
    \resizebox{0.95\textwidth}{!}{
	\begin{tabular}{c|cc|cc|c|c}

		\specialrule{1pt}{0pt}{2pt}   
		\specialrule{0.3pt}{0pt}{0pt} 
		
		\multirow{3}{*}{Method} 
        & \multicolumn{5}{c|}{7-Scenes - 1000 Frames} 
        & \multirow{3}{*}{Time $\downarrow$} \\ \cline{2-6}
        
        & \multicolumn{2}{c|}{Point Cloud Estimation}
        & \multicolumn{2}{c|}{Camera Pose Estimation}
        & \multicolumn{1}{c|}{Depth Estimation}
        & \\ \cline{2-6}
        
		& Acc$\downarrow$ 
        & NC$\uparrow$ 
        & RRA$\uparrow$ 
        & RTA$\uparrow$ 
        & $\delta{<}1.25$$\uparrow$ 
        & \\
        \hline
        
        SparseVGGT 
        & 0.022 & 0.616 & 99.82 & 94.38 & 0.956 & 113.5s \\
        
        FastVGGT 
        & 0.019 & 0.623 & \textbf{100.00} & 95.84 & 0.953 & 71.5s \\
        
        TurboVGGT (Ours) 
        & \textbf{0.017} & \textbf{0.628} & \textbf{100.00} 
        & \textbf{96.35} & \textbf{0.979} & \textbf{55.4s} \\

        \specialrule{1pt}{0pt}{2pt}   
		\specialrule{0.3pt}{0pt}{0pt} 
		
		\multirow{3}{*}{Method} 
        & \multicolumn{5}{c|}{7-Scenes - 500 Frames} 
        & \multirow{3}{*}{Time $\downarrow$} \\ \cline{2-6}
        
        & \multicolumn{2}{c|}{Point Cloud Estimation}
        & \multicolumn{2}{c|}{Camera Pose Estimation}
        & \multicolumn{1}{c|}{Depth Estimation}
        & \\ \cline{2-6}
        
		& Acc$\downarrow$ 
        & NC$\uparrow$ 
        & RRA$\uparrow$ 
        & RTA$\uparrow$ 
        & $\delta{<}1.25$$\uparrow$ 
        & \\
        \hline
        
        SparseVGGT 
        & 0.021 & 0.620 & 99.94 & 94.90 & 0.956 & 32.3s \\
        
        FastVGGT 
        & 0.018 & 0.625 & \textbf{100.00} & 96.13 & 0.953 & 27.2s \\
        
        TurboVGGT (Ours) 
        & \textbf{0.017} & \textbf{0.627} & \textbf{100.00} 
        & \textbf{96.66} & \textbf{0.980} & \textbf{17.8s} \\
        
        \hline
	\end{tabular}
    }    
    \label{tab:long_sequence_7scenes}
\end{table}

\subsection{Supplementary Visualization}
\begin{figure}[h]
    \centering
    \includegraphics[width=0.99\linewidth]{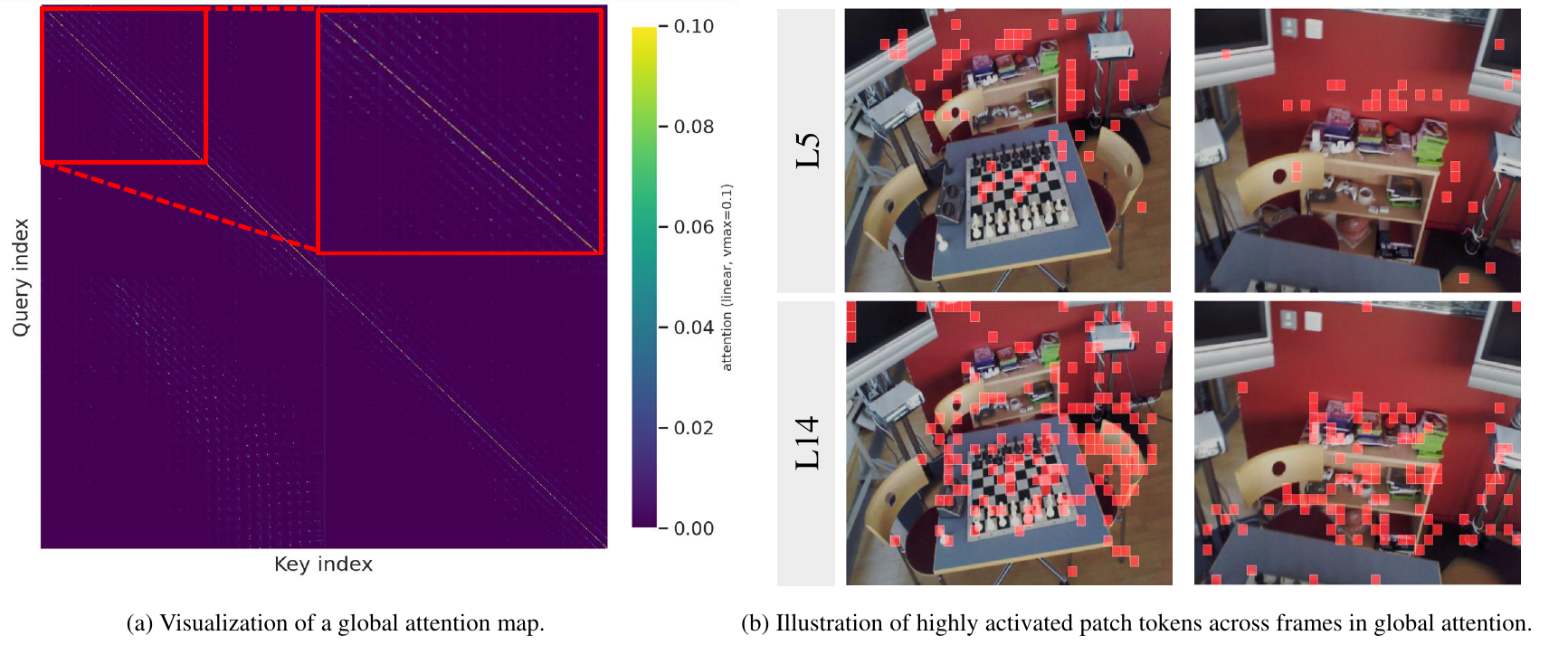}
    \caption{Supplementary visualization of the motivation for the proposed adaptive alternating attention (Enlarged visualizations).
    (a) A visualization of an average global attention map in the alternating attention block.
    (b) An illustration of highly activated patch tokens of different frames in global attention.
    }
    \label{fig:suppl_motivation}
\end{figure}

In Figure~\ref{fig:suppl_motivation}, we provide supplementary visualizations of the motivation for our proposed adaptive alternating attention, which are enlarged visualizations of Figure~\ref{fig:motivation} in the main paper.

\subsection{More Qualitative Results}
\begin{figure}[h]
    \centering
    \includegraphics[width=0.99\linewidth]{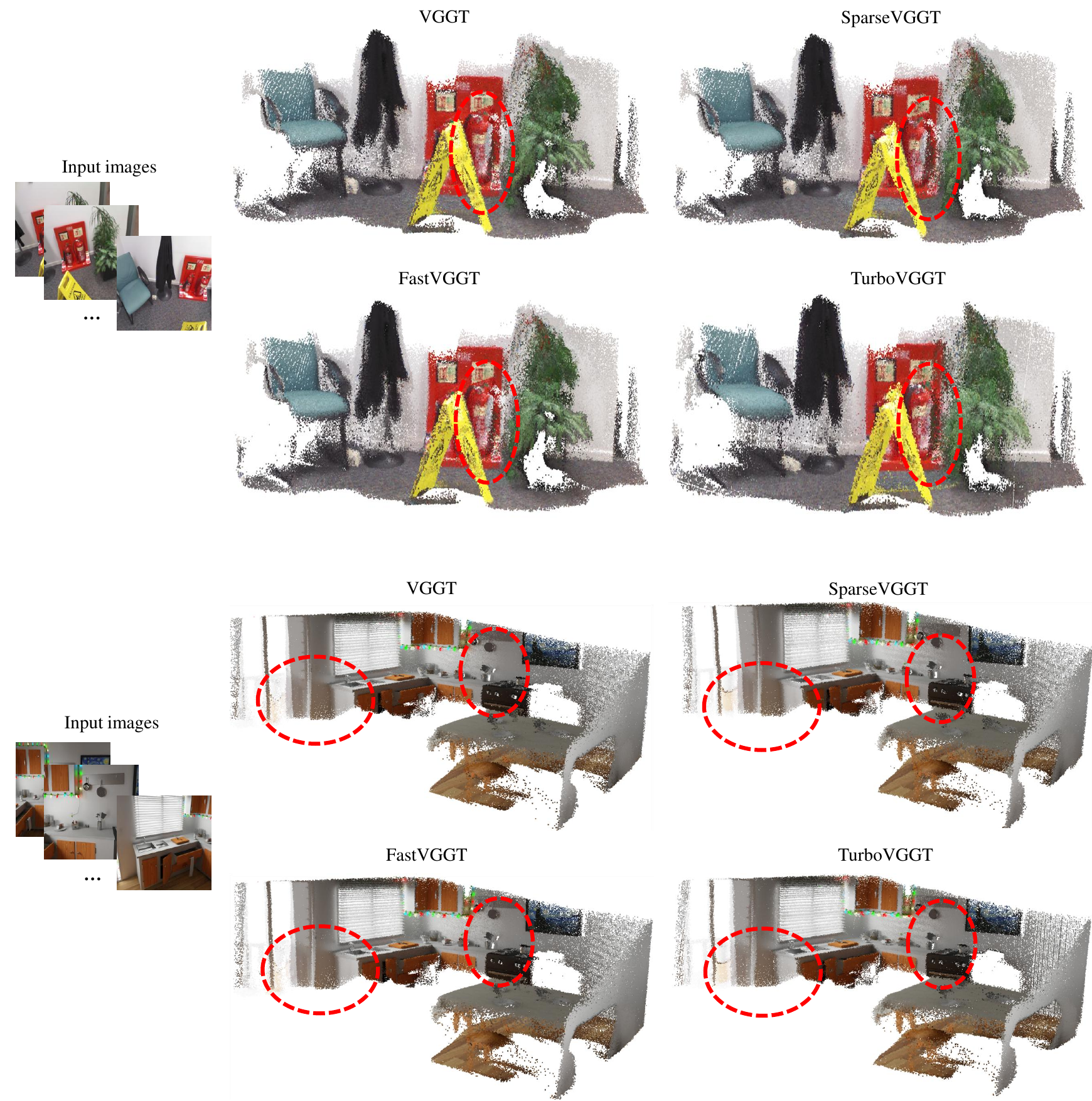}
    \caption{Qualitative comparison of point cloud reconstruction.}
    \label{fig:suppl_vis_point}
\end{figure}

\begin{figure}[h]
    \centering
    \includegraphics[width=0.99\linewidth]{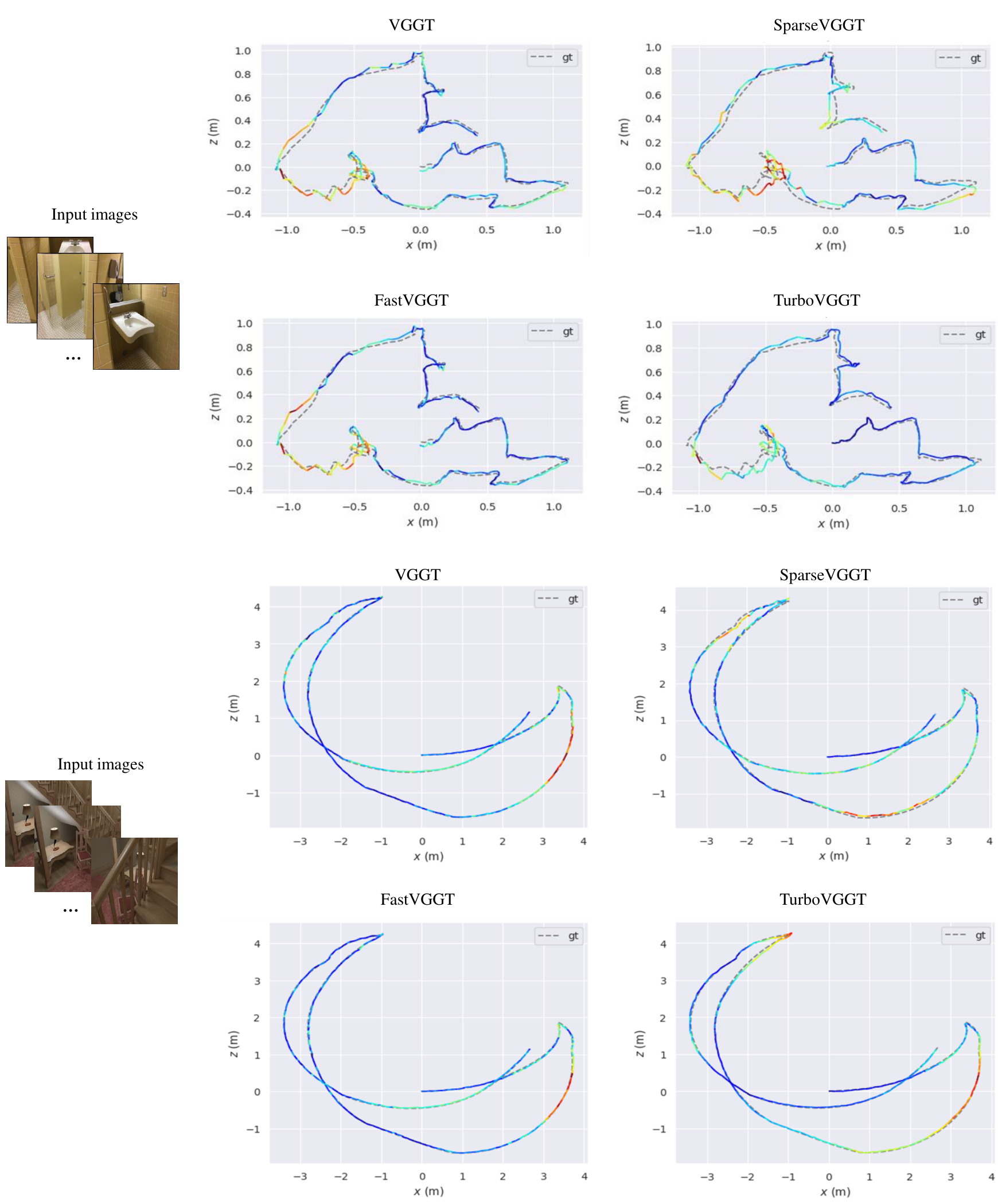}
    \caption{Qualitative comparison of camera pose estimation.}
    \label{fig:suppl_vis_camera}
\end{figure}

\begin{figure}[h]
    \centering
    \includegraphics[width=0.99\linewidth]{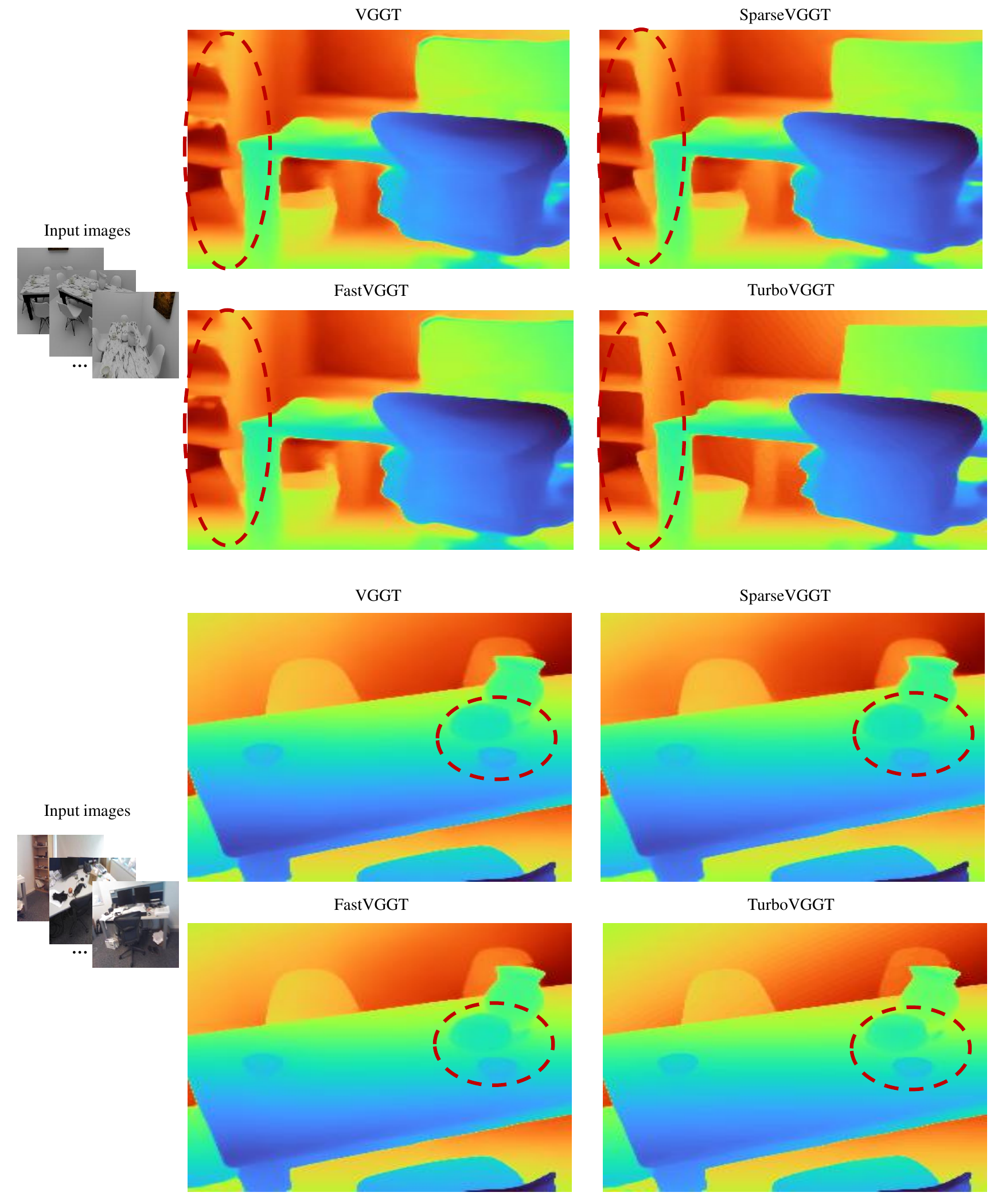}
    \caption{Qualitative comparison of depth estimation.}
    \label{fig:suppl_vis_depth}
\end{figure}

In Figures~\ref{fig:suppl_vis_point}, \ref{fig:suppl_vis_camera}, and \ref{fig:suppl_vis_depth}, we provide more qualitative results of point cloud reconstruction, camera pose estimation, and depth estimation.
From these results, we can see that our TurboVGGT can effectively reconstruct point clouds, estimate camera poses, and predict depth maps.
These results further demonstrate the superior performance of our TurboVGGT compared with state-of-the-art methods.

\clearpage
\section*{NeurIPS Paper Checklist}

\begin{enumerate}

\item {\bf Claims}
    \item[] Question: Do the main claims made in the abstract and introduction accurately reflect the paper's contributions and scope?
    \item[] Answer: \answerYes{}.
    \item[] Justification: See the Abstract section and the Introduction section of the paper.

\item {\bf Limitations}
    \item[] Question: Does the paper discuss the limitations of the work performed by the authors?
    \item[] Answer: \answerYes{}.
    \item[] Justification: See the limitation discussion in the Technical Appendices and Supplementary Material section of the paper.

\item {\bf Theory assumptions and proofs}
    \item[] Question: For each theoretical result, does the paper provide the full set of assumptions and a complete (and correct) proof?
    \item[] Answer: \answerNA{}.
    \item[] Justification: This work does not include theoretical results, so this question is not applicable.

    \item {\bf Experimental result reproducibility}
    \item[] Question: Does the paper fully disclose all the information needed to reproduce the main experimental results of the paper to the extent that it affects the main claims and/or conclusions of the paper (regardless of whether the code and data are provided or not)?
    \item[] Answer: \answerYes{}.
    \item[] Justification: We provide detailed descriptions of the techniques in the Methodology section and the Experiments section of the paper.

\item {\bf Open access to data and code}
    \item[] Question: Does the paper provide open access to the data and code, with sufficient instructions to faithfully reproduce the main experimental results, as described in supplemental material?
    \item[] Answer: \answerNo{}.
    \item[] Justification: This work is built on top of publicly available codebases and uses publicly available datasets, and technical details of the proposed approach are described in the Methodology section and the Experiments section of the paper.

\item {\bf Experimental setting/details}
    \item[] Question: Does the paper specify all the training and test details (e.g., data splits, hyperparameters, how they were chosen, type of optimizer) necessary to understand the results?
    \item[] Answer: \answerYes{}.
    \item[] Justification: See the Experiments section of the paper.

\item {\bf Experiment statistical significance}
    \item[] Question: Does the paper report error bars suitably and correctly defined or other appropriate information about the statistical significance of the experiments?
    \item[] Answer: \answerNo{}.
    \item[] Justification: It is not a common practice in this area to report statistical significance. Following existing work, we evaluate our approach using widely adopted benchmarks and evaluation metrics, and our results do not change much across different runs.

\item {\bf Experiments compute resources}
    \item[] Question: For each experiment, does the paper provide sufficient information on the computer resources (type of compute workers, memory, time of execution) needed to reproduce the experiments?
    \item[] Answer: \answerYes{}.
    \item[] Justification: See the Experiments section of the paper.
    
\item {\bf Code of ethics}
    \item[] Question: Does the research conducted in the paper conform, in every respect, with the NeurIPS Code of Ethics \url{https://neurips.cc/public/EthicsGuidelines}?
    \item[] Answer: \answerYes{}.
    \item[] Justification: This work conforms with the NeurIPS Code of Ethics.

\item {\bf Broader impacts}
    \item[] Question: Does the paper discuss both potential positive societal impacts and negative societal impacts of the work performed?
    \item[] Answer: \answerYes{}.
    \item[] Justification: See the potential positive and negative societal impacts discussed in the Technical Appendices and Supplementary Material section of the paper.
    
\item {\bf Safeguards}
    \item[] Question: Does the paper describe safeguards that have been put in place for responsible release of data or models that have a high risk for misuse (e.g., pre-trained language models, image generators, or scraped datasets)?
    \item[] Answer: \answerNA{}.
    \item[] Justification: This work poses no such risks.

\item {\bf Licenses for existing assets}
    \item[] Question: Are the creators or original owners of assets (e.g., code, data, models), used in the paper, properly credited and are the license and terms of use explicitly mentioned and properly respected?
    \item[] Answer: \answerYes{}.
    \item[] Justification: This work has properly cited the original papers and followed the licenses.

\item {\bf New assets}
    \item[] Question: Are new assets introduced in the paper well documented and is the documentation provided alongside the assets?
    \item[] Answer: \answerNA{}.
    \item[] Justification: This work does not release new assets.

\item {\bf Crowdsourcing and research with human subjects}
    \item[] Question: For crowdsourcing experiments and research with human subjects, does the paper include the full text of instructions given to participants and screenshots, if applicable, as well as details about compensation (if any)? 
    \item[] Answer: \answerNA{}.
    \item[] Justification: This work does not involve crowdsourcing nor research with human subjects.

\item {\bf Institutional review board (IRB) approvals or equivalent for research with human subjects}
    \item[] Question: Does the paper describe potential risks incurred by study participants, whether such risks were disclosed to the subjects, and whether Institutional Review Board (IRB) approvals (or an equivalent approval/review based on the requirements of your country or institution) were obtained?
    \item[] Answer: \answerNA{}.
    \item[] Justification: This work does not involve crowdsourcing nor research with human subjects.

\item {\bf Declaration of LLM usage}
    \item[] Question: Does the paper describe the usage of LLMs if it is an important, original, or non-standard component of the core methods in this research? Note that if the LLM is used only for writing, editing, or formatting purposes and does \emph{not} impact the core methodology, scientific rigor, or originality of the research, declaration is not required.
    \item[] Answer: \answerNA{}.
    \item[] Justification: LLMs are not an important, original, or non-standard component of the core methods in this research.

\end{enumerate}

\end{document}